%% file: aaai22.tex
\def\year{2022}\relax
\documentclass[letterpaper]{article} 
\usepackage{aaai22}  
\usepackage{times}  
\usepackage{helvet}  
\usepackage{courier}  
\usepackage[hyphens]{url}  
\usepackage{graphicx} 
\usepackage{natbib}  
\usepackage{caption} 
\DeclareCaptionStyle{ruled}{labelfont=normalfont,labelsep=colon,strut=off} 
\usepackage{algorithm}
\usepackage{algorithmic}

\usepackage{newfloat}
\usepackage{listings}
\floatstyle{ruled}
\newfloat{listing}{tb}{lst}{}
\floatname{listing}{Listing}
\usepackage[table]{xcolor}
\usepackage{soul}
\colorlet{soulred}{orange!35}
\sethlcolor{soulred}
\usepackage{subcaption}
\usepackage{array,multirow}

\usepackage{amsmath}
\usepackage{amssymb}

\title{Flexible Instance-Specific Rationalization of NLP Models}
\author {
    George Chrysostomou, 
    Nikolaos Aletras 
}
\affiliations {
    Department of Computer Science, University of Sheffield   
    
    gchrysostomou1@sheffield.ac.uk, n.aletras@sheffield.ac.uk
}

\usepackage{bibentry}

\begin{document}

\maketitle

\begin{abstract}
Recent research on model interpretability in natural language processing extensively uses feature scoring methods for identifying which parts of the input are the most important for a model to make a prediction (i.e. explanation or rationale). 
However, previous research has shown that there is no clear best scoring method across various text classification tasks while practitioners typically have to make several other ad-hoc choices regarding the length and the type of the rationale (e.g. short or long, contiguous or not).
Inspired by this, we propose a simple yet effective and flexible method that allows selecting optimally for each data instance: (1) a feature scoring method; (2) the length; and (3) the type of the rationale.
Our method is inspired by input erasure approaches to interpretability which assume that the most faithful rationale for a prediction should be the one with the highest difference between the model's output distribution using the full text and the text after removing the rationale as input respectively. 
Evaluation on four standard text classification datasets shows that our proposed method provides more faithful, comprehensive and highly sufficient explanations compared to using a fixed feature scoring method, rationale length and type. More importantly, we demonstrate that a practitioner is not required to make any ad-hoc choices in order to extract faithful rationales using our approach.\footnote{Code for experiments available at: \url{https://github.com/GChrysostomou/instance-specific-rationale}} 

\end{abstract}

\section{Introduction}

Large pre-trained transformer-based language models such as \texttt{BERT} \cite{devlin-etal-2019-bert, bommasani2021opportunities}, currently dominate performance across language understanding benchmarks~\cite{wang2018glue}. These developments have opened up new challenges on how to extract faithful explanations (i.e. rationales\footnote{We use these terms interchangeably throughout the paper.}), which accurately represent the true reasons behind a model's prediction when adapted to downstream tasks \cite{jacovi-goldberg-2020-towards}.

Recent studies use feature scoring (i.e. attribution) methods such as gradient and attention-based scores \cite{larras2016explaining, integrated_gradients, jain2019attention, chrysostomou-aletras-2021-improving} to identify important (i.e. salient) segments of the input to subsequently extract them as rationales \cite{jain2020learning, treviso-martins-2020-explanation}. 
However, a single feature scoring method is typically applied across the whole dataset (i.e. globally). This might not be optimal for individual instances resulting into less faithful explanations~\cite{jacovi-goldberg-2020-towards,atanasova2020diagnostic}. 
Additionally, rationales are usually extracted using a pre-defined fixed length (i.e. the ratio of a rationale compared to the full input sequence) and type (i.e. top $k$ terms or contiguous) globally. We hypothesize that using a fixed length or type for different instances could result into shorter (i.e. not sufficient for explaining a model's prediction) or longer than needed rationales reducing rationale faithfulness, whilst finding the explanation length is an open problem \citep{zhang-etal-2021-sample}.
Moreover to extract rationales, practitioners are currently required to make assumptions for the rationale parameters (i.e. feature scoring method, length and type), whilst different choice of parameters might substantially affect the faithfulness of the rationales.

In this paper, we propose a simple yet effective method that operates at instance-level and mitigates the a priori selection of a specific (1) feature scoring method; (2) length and (3) type when extracting faithful rationales. Our proposed method is flexible and allows the automatic selection of some of these instance-specific parameters or all. Inspired by erasure methods, it functions by computing differences between a model's output distributions obtained using the full input sequence and the input without the rationale respectively. We base this on the assumption that by removing important tokens from the sequence, we should observe large divergences in the model's predicted distribution \cite{nguyen-2018-comparing, serrano2019attention, deyoung-etal-2020-eraser} resulting into more faithful rationales~\cite{atanasova2020diagnostic, chen-ji-2020-learning}. 
The contributions of our work are thus as follows:


\begin{itemize}

    \item To the best of our knowledge, we are the first to propose a method for instance-specific faithful rationale extraction;
    
    \item We empirically demonstrate that rationales extracted with instance-specific flexible feature scoring method, length and type using our proposed method are more comprehensive than rationales with fixed, pre-defined parameters;
    
    \item We show that our method results in consistently highly sufficient rationales, mitigating the variability in faithfulness of different feature scoring methods across datasets when used globally, i.e. the same for all instances~\citep{atanasova2020diagnostic}.
    


\end{itemize}

\section{Background and Related Work}

\subsection{Rationale Extraction
\label{sec:extracting_rationales}}

Given a trained model $\mathcal{M}$, an input $\mathbf{x} = [x_1, \hdots, x_T]$ and a predicted distribution over classes  $\mathcal{Y}$, rationale extraction methods seek to identify the most important subset $\mathcal{R} \in  \mathbf{x}$ of the input for explaining the model's prediction. 


There are two common approaches for extracting rationales. The first consists of two modules jointly trained on an end-task, e.g. sentiment analysis \cite{lei-etal-2016-rationalizing, bastings-etal-2019-interpretable}. The first module extracts the rationale (i.e. typically by learning to select which inputs should be masked) and the second module is trained using only the rationale. The second approach consists of using feature scoring (or attribution) methods (i.e. salience metrics) to first identify important parts of the input and then extract the rationales from $\mathcal{M}$~\cite{jain2020learning, treviso-martins-2020-explanation, deyoung-etal-2020-eraser}. 
A limitation of the first approach is that the models are hard to train compared to the latter and often do not reach high accuracy \cite{jain2020learning}. Regarding the latter approach, a limitation is that the same feature scoring method is applied to all instances in a given dataset, irrespective of whether a feature scoring method is not the best for a particular instance \citep{atanasova2020diagnostic, jacovi-goldberg-2020-towards} while finding a suitable explanation length is an open problem \citep{zhang-etal-2021-sample}.

\subsection{Computing Input Importance}

Feature scoring methods $\Omega$ compute input importance scores $\boldsymbol{\omega}$ for each token in the sequence $\mathbf{x}$, such that $\boldsymbol{\omega} = \Omega(\mathcal{M}, \mathbf{x}, \mathcal{Y})$. High scores indicate that the associated tokens contributed more towards a model's prediction. Subsequently, $\mathcal{R}$ is extracted by selecting the $K$ highest scored tokens (or $K$-gram for contiguous) in a sequence \citep{deyoung-etal-2020-eraser, jain2020learning}.

A common approach to computing $\boldsymbol{\omega}$ is by calculating the gradients of the prediction with respect to the input~\cite{kindermans2016investigating,li-etal-2016-visualizing,larras2016explaining, integrated_gradients, bastings-filippova-2020-elephant}. \citet{jain2020learning} use attention weights to attribute token importance for rationale extraction, while \citet{treviso-martins-2020-explanation} propose sparse attention. \citet{li-etal-2016-visualizing} compute input importance scores by measuring the difference in a model's prediction between keeping and omitting each token, with \citet{kim-etal-2020-interpretation} also suggesting input marginalization as an alternative to token omission. Another way is using sparse linear meta-models that are easier to interpret~\cite{ribeiro2016model}. \citet{atanasova2020diagnostic} however show that sparse linear meta-models are not as faithful as gradient-based approaches for interpreting large language models.

\subsection{Evaluating Rationale Faithfulness 
\label{sec:evaluating_expl}}

Having extracted $\mathcal{R}$, we typically need to evaluate how faithful that explanation is for a model's prediction. Several studies evaluate the faithfulness of explanations by training a separate classifier on an end-task using only the rationales as input \cite{jain2020learning, treviso-martins-2020-explanation}. 
These classifiers are inherently faithful, as they are trained only on the rationales \cite{jain2020learning}. %
Other studies compare the ability of different feature scoring methods to identify important tokens by using word erasure, i.e. masking~\cite{samek_et_al, serrano2019attention, atanasova2020diagnostic, chen-ji-2020-learning, deyoung-etal-2020-eraser, zhang-etal-2021-sample, chrysostomou-aletras-2021-enjoy}. The intuition is that by removing the most important tokens, it should result in a larger difference in the output probabilities, compared to removing a less important token which will also lead to drops in classification accuracy \cite{robnik2008explaining, nguyen-2018-comparing, atanasova2020diagnostic}. \citet{deyoung-etal-2020-eraser} use erasure to evaluate the comprehensiveness and sufficiency of rationales. \citet{carton-etal-2020-evaluating} suggest normalizing these metrics using the predictions of the model with a baseline input, to allow for a fairer comparison across models and datasets.

\section{Instance-Specific Rationale Extraction
\label{sec:methodology}}

Our aim is to address the \emph{``one-size-fits-all''} ad-hoc approach of previous work on rationale extraction with feature scoring methods that typically extracts rationales using the same feature scoring method, length and type across all instances in a dataset. Inspired by word erasure approaches \cite{nguyen-2018-comparing, serrano2019attention, deyoung-etal-2020-eraser} we mask the tokens that constitute a rationale and record the difference $\delta$ in a model's output distribution by using the full text and the reduced input. Our main assumption is that a sufficiently faithful rationale is the one that will result into the largest $\delta$~\cite{atanasova2020diagnostic, chen-ji-2020-learning, deyoung-etal-2020-eraser}. 
Following this assumption, we can extract rationales by selecting for each instance a specific (1) \emph{feature scoring method}; (2) \emph{length}; and (3) \emph{type}.\footnote{Similar to \citet{jain2020learning}, we consider two rationale types: (a) \textsc{TopK} tokens ranked by a feature scoring method, treating each word in the input sequence independently; and (b) \textsc{Contiguous} span of input tokens of length \emph{K} with the highest overall score computed by a feature scoring method.}

\subsection{Instance-level Feature Scoring Selection}

Given a set of $M$ feature scoring methods $\{\Omega_1, \hdots ,\Omega_M\}$, we extract a rationale $\mathcal{R}$ as follows:

\begin{enumerate}
    \item For each $\Omega_i$ in the set we compute input importance scores $\boldsymbol{\omega}_i = \Omega_i(\mathcal{M}, \mathbf{x}, \mathcal{Y})$;
    
    \item We subsequently select the $K$ highest scored tokens (\textsc{TopK}) or the highest $K$-gram (\textsc{Contiguous)} to form a rationale $\mathcal{R}_i$, where $K$ is the rationale length; 
    \item For each rationale we compute the difference $\delta_i$, between the reference model output (using full text input) and the model output having masked the rationale, such that:
    $$\delta_i = \Delta(\mathcal{Y}, \mathcal{Y}^m_i)= \Delta(\mathcal{M}(\mathbf{x})  , \mathcal{M}(\mathbf{x}_{\backslash \mathcal{R}_i}))$$
    where $\Delta$ is the function used to compute the difference between the two outputs;
    \item We select the rationale $\mathcal{R}$ with the highest difference $\delta_{max} = max(\{\delta_1, \hdots, \delta_i, \hdots, \delta_M\})$.
\end{enumerate}

For computing $\delta$, we experiment with the following divergence metrics ($\Delta$): (a) Kullback-Leibler (\textsc{KL}); (b) Jensen-Shannon divergence (\textsc{JSD}); (c) Perplexity (\textsc{Perp.}) and (d) Predicted Class Probability (\textsc{ClassDiff}).\footnote{We describe the metrics in detail in App. \ref{appendix:divergence_metrics}.}

\subsection{Instance-level Rationale Length Selection}
\label{ref:text}

For computing at instance-level the rationale length $k$ and extracting the rationale $R$ using a single feature scoring method $\Omega$, we propose the following steps:

\begin{enumerate}
    \item Given $\Omega$, we first compute input importance scores $\boldsymbol{\omega} = \Omega(\mathcal{M}, \mathbf{x}, \mathcal{Y})$;
    \item We then iterate over the sequence such that $k = range(1 , N)$, where $N$ is the fixed, pre-defined rationale length and $k$ the possible rationale length at the current iteration. We set $N$ as the upper bound rationale length for our approach to make results comparable with fixed length rationales.
    \item At each iteration we begin by masking the top $k$ tokens (as indicated by $\omega$) to form a candidate rationale $\mathcal{R}_{k}$. When using \textsc{TopK} we mask the $k$ highest scored tokens, whilst with \textsc{Contiguous} we mask the highest scored $k$-gram;
    \item We compute the difference $\delta_{k}$ between the reference model output $\mathcal{Y}$ and the model output having masked the candidate rationale $\mathcal{Y}^m_k = \mathcal{M}(\mathbf{x}_{\backslash \mathcal{R}_{k}})$;
    \item We record every $\delta$ until $k=N$ and extract the rationale $\mathcal{R}$ with the highest difference $\delta_{max} = max(\{\delta_{1}, \hdots, \delta_{k}, \hdots, \delta_{N}\})$, where $k$ at $\delta_{max}$ is the computed rationale length.\footnote{We also experimented with early stopping, whereby the difference between $\delta_k$ and the $\delta_{max}$ until $k$ are under a specified threshold, however this resulted in reduced performance (we included a more thorough analysis in the App. \ref{appendix:alternative_delta}).}
\end{enumerate}

\subsection{Instance-level Rationale Type Selection}

In a similar way to selecting a feature scoring method, our approach can also be used to select between different rationale types (i.e. \textsc{Contiguous} or \textsc{TopK}) for each instance in the dataset. 

Finally, our approach is flexible and can be easily modified to support selecting any of these parameters while keeping the rest fixed (i.e. feature scoring method, rationale length and rationale type) or by selecting any combination of them. An important benefit of our approach is that we extract rationales with different settings for each instance rather than using uniform settings globally (i.e. across the whole dataset), which we empirically demonstrate to be beneficial for faithfulness (\S \ref{sec:results}).

\section{Experimental Setup}

\subsection{Tasks}

For our experiments we use the following datasets (details in Table \ref{tab:data_characteristics}):

\begin{itemize}
    \item \textbf{SST: }Binary sentiment classification without neutral sentences \cite{socher-etal-2013-recursive}.
    \item \textbf{AG: }News articles categorized in Science, Sports, Business, and World topics \cite{del2005ranking}.
    \item \textbf{Evidence Inference \textsc{(Ev.Inf.)}: }Abstract-only biomedical articles  describing  randomized  controlled trials.  The task is to infer the relationship between a given intervention and comparator with respect to an outcome \cite{lehman-etal-2019-inferring}.
    \item \textbf{MultiRC (\textsc{M.Rc}): }A reading comprehension task with questions having multiple correct answers that depend on information from multiple sentences \cite{khashabi-etal-2018-looking}. Following \citet{deyoung-etal-2020-eraser} and \citet{jain2020learning}, we convert this to a binary classification task where each rationale/question/answer triplet forms an instance and each candidate answer is labeled as True/False
\end{itemize}

\subsection{Models}

Similar to \citet{jain2020learning}, we use \textsc{BERT} \cite{devlin-etal-2019-bert} for \textsc{SST} and \textsc{AG}); \textsc{SciBERT} \cite{beltagy-etal-2019-scibert} for \textsc{Ev.Inf.} and \textsc{Roberta} \cite{roberta_paper} for \textsc{M.RC}. See App. \ref{appendix:model_hyperparameters} for hyperparameters.


\begin{table}[!t]
\setlength\tabcolsep{2pt}
\small
\centering
\begin{tabular}{l||ccccc}
\textbf{Data}   & $|W|$ & \textbf{C} & \textbf{\begin{tabular}[c]{@{}c@{}}Splits\\   Train/Dev/Test\end{tabular}} & F1 & $N$\\ \hline
\textbf{SST}                & 18                  & 2     & 6,920 / 872 / 1,821  
 & 90.1 $\pm$ 0.2 & 20\% \\
 
\textbf{AG}         & 36                  & 4   & 102,000 / 18,000 / 7,600         & 93.5 $\pm$ 0.2 & 20\% \\
\textbf{Ev.Inf.}               & 363                 & 3     & 5,789 / 684 / 720   & 83.0 $\pm$ 1.6 & 10\% \\

\textbf{M.RC}             & 305                  & 2     & 24,029 / 3,214 / 4,848    & 73.2 $\pm$ 1.7 & 20\% \\
\end{tabular}
\caption{Dataset statistics including average words at instance ($|W|$), number of classes (\textbf{C}), data splits, F1 macro performance and the fixed, pre-defined rationale ratio across all instances ($N$). }
\label{tab:data_characteristics}
\end{table}

\subsection{Feature Scoring Methods 
\label{sec:feature_attribution}}

We use a random baseline and six other feature scoring methods (to compute input importance scores) similar to \citet{jain2020learning} and \citet{serrano2019attention}.

\begin{itemize}
\item \textbf{Random (\textsc{Rand}): } Random allocation of token importance. 
\item\textbf{Attention ($\boldsymbol{\alpha}$):}  Token importance corresponding to normalized attention scores \cite{jain2020learning}.
\item\textbf{Scaled Attention ($\boldsymbol{\alpha} \nabla \boldsymbol{\alpha}$):} Scales the attention scores $\alpha_i$ with their corresponding gradients $\nabla \alpha_i= \frac{\partial \hat{y}}{\partial \alpha_i}$ \cite{serrano2019attention} .
\item\textbf{InputXGrad ($\mathbf{x} \nabla \mathbf{x}$):} Attributes input importance by multiplying the gradient of the input by the input with respect to the predicted class, where $\nabla x_i = \frac{\partial \hat{y}}{\partial x_i}$ \cite{kindermans2016investigating, atanasova2020diagnostic} .

\item\textbf{\textbf{Integrated Gradients ($\mathbf{IG}$)}:} Ranking words by computing the integral of the gradients taken along a straight path from a baseline input (zero embedding vector) to the original input \cite{integrated_gradients}. 

\item\textbf{\textbf{DeepLift}:} Ranking words according to the difference between the activation of each neuron to a reference activation \cite{shrikumar-deeplift}. 

\item\textbf{\textbf{LIME}:} Ranking words by learning an interpretable
model locally around the prediction \cite{ribeiro2016model}.
\end{itemize}
\subsection{Evaluating Explanation Faithfulness
\label{sec:evaluating_expl_faith_meth}}

\begin{itemize}
    \item \textbf{F1 macro: }Similar to \citet{arras2017explaining} we measure the F1 macro performance of model $\mathcal{M}$ when masking the rationale in the original input ($\mathbf{x}_{\setminus \mathcal{R}}$). A key difference in our approach is that we use the predicted labels of the model with full input as gold labels, as we are interested in the faithfulness of explanations for the predictions of the model. Larger drops in F1 scores indicate that the extracted rationale is more faithful.\footnote{We also conducted experiments using the dataset gold labels with results being comparable. We include all results in App. \ref{appendix:results}.}
    
    \item \textbf{Normalized Sufficiency (NormSuff): } We measure the degree to which the extracted rationales are sufficient for a model to make a prediction \cite{deyoung-etal-2020-eraser}. Similar to \citet{carton-etal-2020-evaluating} we bind sufficiency between 0 and 1 and use the reverse difference so that higher is better. We modify this metric and measure the normalized sufficiency \cite{carton-etal-2020-evaluating} such that: 

    \begin{equation}
    \begin{aligned}
        \text{Suff}(\mathbf{x}, \hat{y}, \mathcal{R}) = 1 - max(0, p(\hat{y}| \mathbf{x})- p (\hat{y}|\mathcal{R}))\\
        \text{NormSuff}(\mathbf{x}, \hat{y}, \mathcal{R}) = \frac{\text{Suff}(\mathbf{x}, \hat{y}, \mathcal{R}) - \text{Suff}(\mathbf{x}, \hat{y}, 0)}{1 - \text{Suff}(\mathbf{x}, \hat{y}, 0)}
    \end{aligned}
    \end{equation}

    \noindent where $\text{Suff}(\mathbf{x}, \hat{y}, 0)$ is the sufficiency of a baseline input (zeroed out sequence) and $\hat{y}$ the model predicted class using the full text $\mathbf{x}$ as input, such that $\hat{y} = \text{arg max}(\mathcal{Y})$. 
    
    \item \textbf{Normalized Comprehensiveness (NormComp): } We measure the extent to which a rationale is needed for a prediction \cite{deyoung-etal-2020-eraser}.  For an explanation to be highly comprehensive, the model's prediction when masking the rationale should have a high difference between the model's prediction with full text. Similarly to \citet{carton-etal-2020-evaluating} we bind this metric between 0 and 1 and normalize it. We compute it by: 

    \begin{equation}
        \begin{aligned}
        \text{Comp}(\mathbf{x}, \hat{y}, \mathcal{R}) = max(0, p(\hat{y}| \mathbf{x})- p (\hat{y}|\mathbf{x}_{\backslash\mathcal{R}}))\\
        \text{NormComp}(\mathbf{x}, \hat{y}, \mathcal{R}) = \frac{\text{Comp}(\mathbf{x}, \hat{y}, \mathcal{R})}{1 - \text{Suff}(\mathbf{x}, \hat{y}, 0)}
        \end{aligned}
    \end{equation}
    
\end{itemize}

We do not conduct human experiments to evaluate explanation faithfulness since that is only relevant to explanation plausibility (i.e. how understandable by humans a rationale is \cite{jacovi-goldberg-2020-towards}) and in practice faithfulness and plausibility do not correlate \cite{atanasova2020diagnostic}. Finally we do not compare with select-then-predict methods \cite{lei-etal-2016-rationalizing, jain2020learning}, as we are interested in faithfully explaining the model $\mathcal{M}$ and not forming inherently faithful classifiers.

\subsection{Performance-Time Trade-off}\label{sec:comput_times}

Input erasure approaches typically require  $N$ forward passes to compute a rationale length (see \S\ref{sec:methodology}) when removing one token at a time. Similar to \citet{nguyen-2018-comparing, atanasova2020diagnostic}, we expedite this process when selecting a rationale length by ``skipping'' every $X$\% of tokens. For our work, we use a 2\% skip rate which led to a seven-fold reduction in the time required to compute rationales for datasets comprising of long sequences, such as MRc and EvInf, with comparable performance in faithfulness to the slower process of removing one token at a time. We include in App. \ref{appendix:complexities} the performance/skip-rate trade-off when (1) we do not use a skip-rate; (2) at 2\% and (3) at 5\%.

\section{Results}\label{sec:results}


\begin{figure*}[!t]
     \centering
     \begin{subfigure}[b]{0.29\textwidth}
         \centering
         \includegraphics[width=\textwidth]{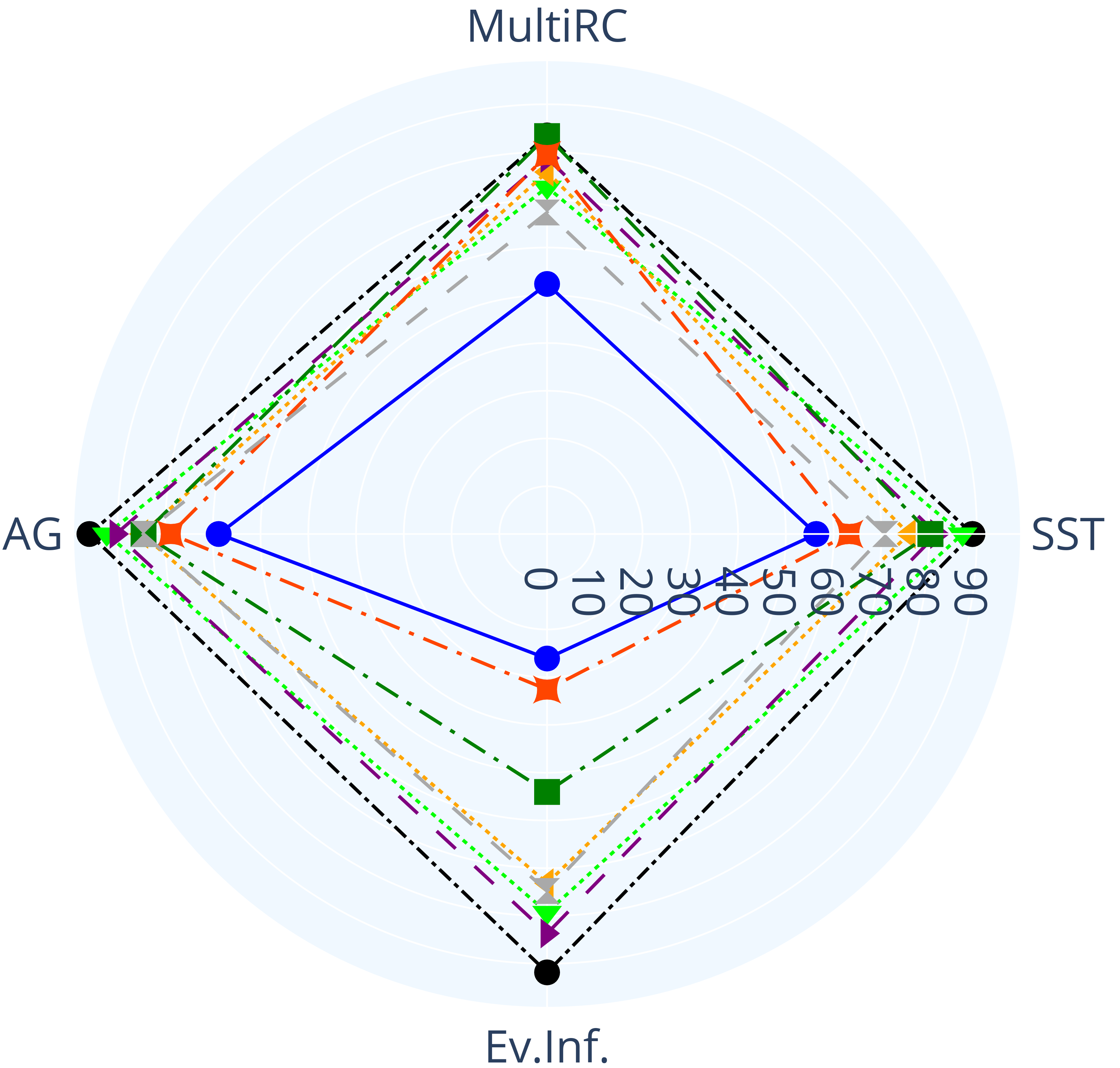}
         \caption{F1 macro}
     \end{subfigure}
     \hspace{0.0em}
     \begin{subfigure}[b]{0.29\textwidth}
         \centering
         \includegraphics[width=\textwidth]{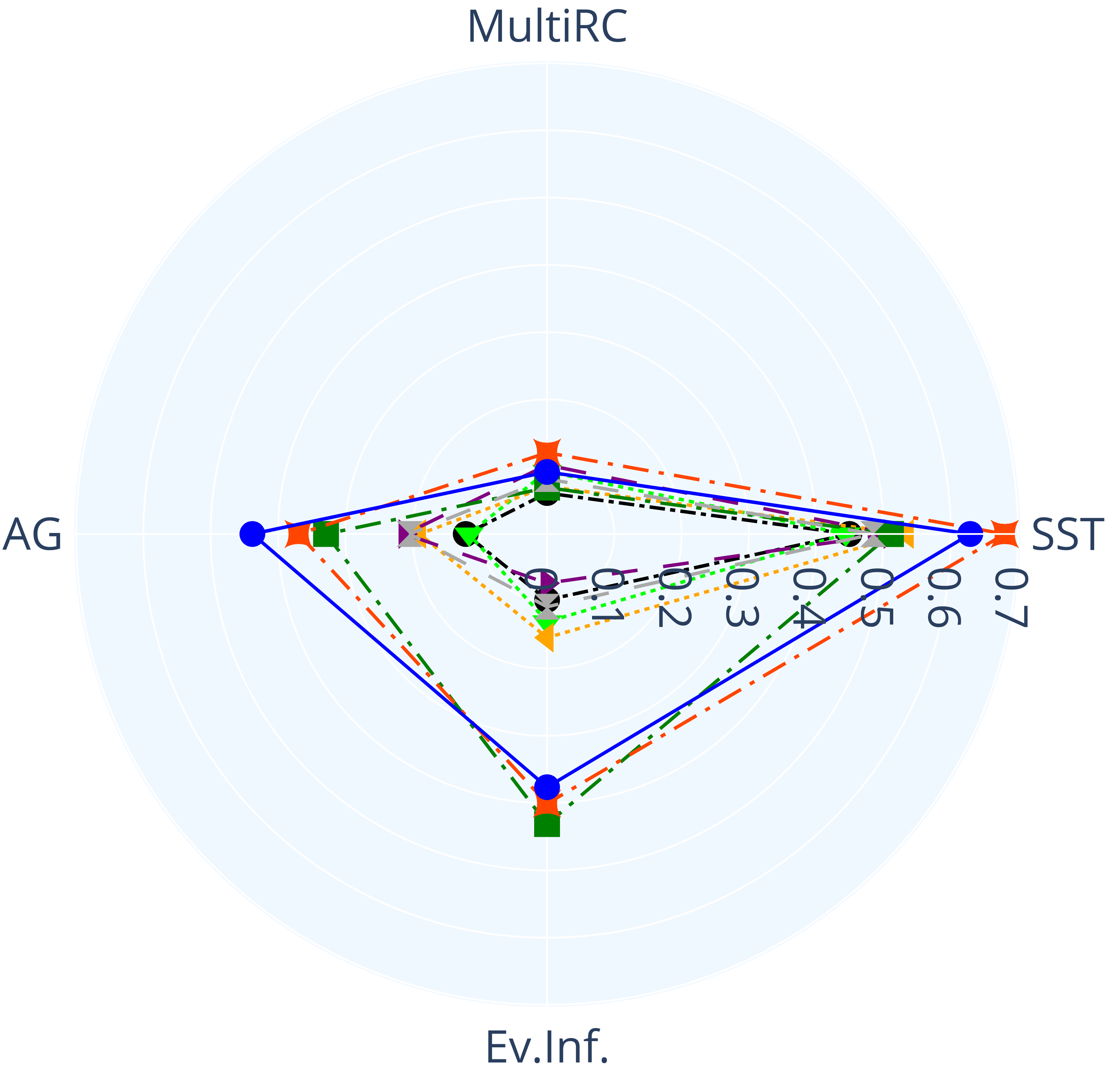}
         \caption{NormSuff}
     \end{subfigure}
     \hspace{0.0em}
     \begin{subfigure}[b]{0.29\textwidth}
         \centering
         \includegraphics[width=\textwidth]{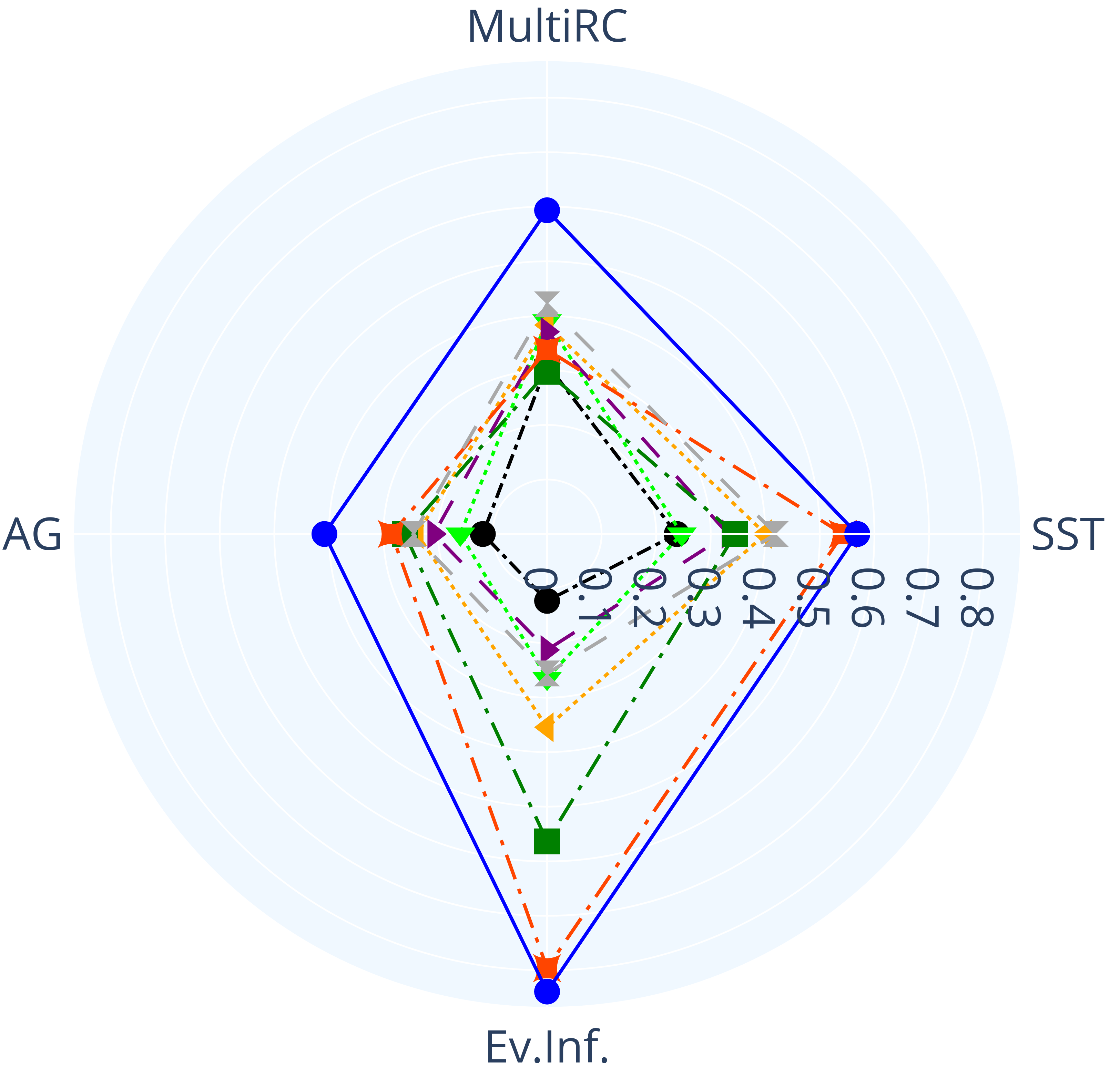}
         \caption{NormComp}
     \end{subfigure}
     \hspace{0.0em}
     \begin{subfigure}[b]{0.09\textwidth}
         \centering
         \includegraphics[width=\textwidth]{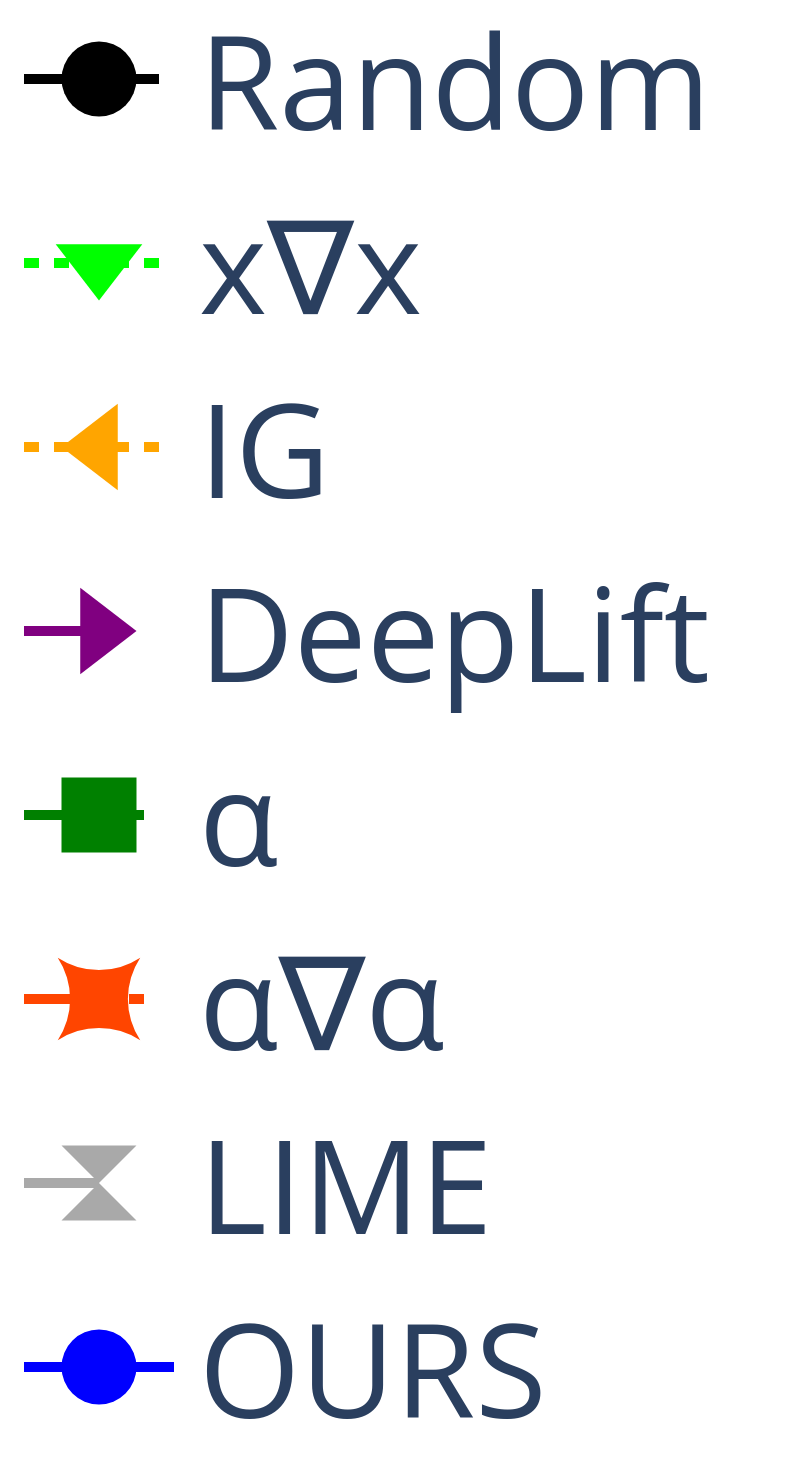}
     \end{subfigure}
        \caption{F1 macro (lower is better), mean NormSuff (higher is better) and mean NormComp (higher is better), when using any single feature scoring method across all instances in a dataset and our proposed method of selecting a feature scoring method for each instance (OURS) for \textsc{TopK} rationale types.}
        
        \label{fig:fixed_len_var_feat}
    
\end{figure*}

\subsection{Selecting Instance-specific Feature Scoring} 
Figure \ref{fig:fixed_len_var_feat} compares the faithfulness of extracted rationales when using our proposed method for selecting an instance-specific feature scoring method (OURS) and our baselines, that use a single fixed pre-defined feature scoring method globally (i.e. across all instances in a dataset). We measure faithfulness using
F1 macro (lower is better), mean NormSuff and mean NormComp (higher is better respectively). For clarity we show results using the \textsc{TopK} rationale type, with results for \textsc{Contiguous} included in App. \ref{appendix:results_feat_scoring_cont}.\footnote{Also for clarity, all results presented in this work are using JSD for $\Delta$. The other divergence functions performed comparably and we include a comparison between them in App. \ref{appendix:comparing_divmetrics}.}

Overall, results demonstrate that rationales extracted with our proposed approach are highly sufficient and comprehensive. In fact, our approach results in more sufficient rationales against all single feature scoring methods in AG and is comparable with the best NormSuff scores in the remainder of the datasets. This suggests that even when rationales with our proposed method are not the most sufficient, they are consistently highly sufficient (i.e. rationales extracted with our approach are significantly more sufficient than fixed, pre-defined feature scoring methods in 18 out of 24 test cases). Compared to our six baselines, the rationales extracted with our approach are significantly more comprehensive across all four datasets (Wilcoxon Rank Sum, $p < .05$). Additionally, the larger drops in F1 macro performance demonstrate that rationales extracted with our proposed approach are more necessary for a model to make a prediction compared to a globally used, pre-defined feature scoring approach. 

Our results strengthen the hypothesis that whilst some feature scoring methods are better than others globally, they might not be optimal for all instances in a dataset \citep{jacovi-goldberg-2020-towards} and our approach helps mitigate that. 
Similar to \citep{atanasova2020diagnostic}, we observe that the faithfulness performance of single feature scoring methods varies across datasets. For example LIME returns more comprehensive rationales than $\alpha\nabla\alpha$ in MultiRC, however is outperformed by the latter in SST. By returning consistently highly comprehensive and sufficient rationales, our propose method helps reducing the variability in faithfulness performance observed when using any single feature scoring method across datasets.

\subsection{Selecting Instance-specific Rationale Length}

Table \ref{tab:direction_of_change} shows the Relative Improvement (R.I.) ratio in mean NormSuff and NormComp ($>$1.0 is better) between rationales extracted using a fixed pre-defined length (see $N$ in Table \ref{tab:data_characteristics}) and rationales extracted using our method with instance-specific length across feature scoring methods and datasets. 
For brevity the detailed results, including F1 macro where we make similar observations to comprehensiveness, are included in App. \ref{appendix:results}.
Overall, rationales extracted using our approach are on average shorter than fixed length rationales. Specifically, rationale length drops from 20\% to 16\% on average in SST, AG; from 20\% to 15\% in M.Rc and from 10\% to 7\% in Ev.Inf..\footnote{We include the computed rationale length results in App. \ref{appx:computed_variable_lengths} and the same evaluation under 2 $\times N$ in App. \ref{appendix:results_double_n}.}

NormSuff scores indicate that our shorter on average rationales are overall less but comparably sufficient with longer, fixed-length rationales. For example with SST rationales with instance-specific length are 0.9-1.0 times less sufficient that rationale with pre-defined length. We find this particularly evident in datasets such as M.Rc and Ev.Inf., where our rationales are on average 4-5\% shorter (approximately 15 tokens shorter on average for $\alpha$ in M.Rc) but still retain comparable sufficiency, while in some cases improving it (e.g. 1.2 R.I. in Ev.Inf. with DeepLift).

We also note that rationales extracted with instance-specific length are more comprehensive in most cases, despite being shorter on average compared to fixed-length rationales. For example in Ev.Inf., \textsc{Contiguous} rationales with I.G. are 1.4 times more comprehensive when we select their length at instance-level. Results also indicate that using our proposed method benefits more \textsc{Contiguous} rationales compared to \textsc{TopK} for comprehensiveness, leading to increased R.I. in the majority of cases. Overall, findings support our initial hypothesis that in certain cases a rationale with longer than needed length might contain unnecessary information and adversely impact its comprehensiveness.

\begin{table}[!t]
\setlength\tabcolsep{1pt}
\small
\centering
\begin{tabular}{ll||cccc||cccc}
  & & \multicolumn{4}{c||}{\textbf{NormSuff}} & \multicolumn{4}{c}{\textbf{NormComp}} \\ 
 & \textsc{\textbf{Feat}}  & \textbf{SST} & \textbf{M.Rc} & \textbf{AG} & \textbf{Ev.Inf.}  & \textbf{SST} & \textbf{M.Rc} & \textbf{AG} & \textbf{Ev.Inf.}  \\\hline
\parbox[t]{3mm}{\multirow{6}{*}{\rotatebox[origin=c]{90}{\textsc{TopK}}}} &  \textbf{DeepLift}                         &         0.9 &             0.8 &        0.8 &             1.1 &               0.8 &                   1.1 &              1.0 &                            1.0 \\

& \textbf{LIME}                             &                  1.0 &             0.7 &        0.9 &                      0.9 &               0.9 &                   1.1 &              1.0 &                   1.0 \\

& $\boldsymbol{\alpha}$                     &         0.9 &             0.9 &        0.7 &             0.8 &               0.8 &                   1.1 &              0.9 &                   1.2 \\

& $\boldsymbol{\alpha\nabla\alpha}$         &         0.9 &             0.9 &        0.8 &                      0.9 &               1.0 &                   1.1 &              0.9 &                   1.0 \\

& \textbf{IG}                               &         0.9 &                      0.9 &        0.8 &                      0.9 &               0.9 &                   1.1 &              1.0 &                   1.1 \\

& $\boldsymbol{\mathbf{x}\nabla\mathbf{x}}$ &         1.0 &             0.8 &        0.7 &                      0.8 &               0.9 &                   1.1 &              0.9 &                   1.2 \\
 \hline\hline

\parbox[t]{3mm}{\multirow{6}{*}{\rotatebox[origin=c]{90}{\textsc{Contiguous}}}} & \textbf{DeepLift}                         &               0.9 &                   0.9 &              0.8 &                   1.2 &                     0.9 &                         1.1 &                             1.3 &                         1.5 \\

& \textbf{LIME}                             &               0.9 &                   0.7 &              0.8 &                            0.9 &                     1.0 &                         1.1 &                             1.2 &                         1.3 \\

& $\boldsymbol{\alpha}$                     &               0.9 &                   0.9 &              0.7 &                   0.9 &                     0.7 &                         1.1 &                    1.0 &                         1.2 \\

& $\boldsymbol{\alpha\nabla\alpha}$         &               0.9 &                   0.8 &              0.8 &                   0.9 &                              1.0 &                         1.1 &                    1.1 &                         1.1 \\

& \textbf{IG}                               &               0.9 &                            0.8 &              0.8 &                            1.0 &                              1.0 &                         1.2 &                             1.2 &                         1.4 \\

& $\boldsymbol{\mathbf{x}\nabla\mathbf{x}}$ &               0.9 &                            0.8 &              0.7 &                            1.0 &                     1.0 &                         1.1 &                    1.0 &                         1.3 \\

\end{tabular}
\caption{Relative Improvement (R.I.) ratios for mean NormSuff and mean NormComp between fixed length rationales (see $N$ in Table \ref{tab:data_characteristics}) extracted using our method and rationales with instance-specific length ($>$1.0 is better).}
\label{tab:direction_of_change}
\end{table}

\begin{table}[!t]
\setlength\tabcolsep{1pt}
\small
\centering
\begin{tabular}{lll||cccc||cccc}
  & & & \multicolumn{4}{c||}{\textbf{NormSuff}} & \multicolumn{4}{c}{\textbf{NormComp}} \\ 
  \textsc{\textbf{Type}} & \textsc{\textbf{Len}} & \textsc{\textbf{Feat}}   & \textbf{SST} & \textbf{M.Rc} & \textbf{AG} & \textbf{Ev.Inf.}  & \textbf{SST} & \textbf{M.Rc} & \textbf{AG} & \textbf{Ev.Inf.}  \\\hline
\parbox[t]{3mm}{\multirow{3}{*}{\rotatebox[origin=c]{90}{\textsc{TopK}}}} & \textsc{Fix} & \textsc{Fix} &        .68 &                      \textbf{.12} &                 .37 &                      .43 &                        .54 &                            .42 &                       .28 &                            .80 \\
& \textsc{I-L} & \textsc{Fix}  &                  .61 &                      .11 &                 .30 &                      .37 &                        .52 &                            .46 &                       .27 &                            .82 \\
& \textsc{Fix} & \textsc{I-L}   &  .63 &                      .09 &                 \textbf{.44} &                      .38 &                        \textbf{.57} &                            .59 &                       \textbf{.41} &                            .84 \\ 
& \textsc{I-L} & \textsc{I-L} &                  .59 &                      .07 &                 .38 &                      .36 &                        .55 &                            .62 &                       .39 &                            .86 \\ \hline

\parbox[t]{3mm}{\multirow{3}{*}{\rotatebox[origin=c]{90}{\textsc{Cont.}}}} & \textsc{Fix} & \textsc{Fix}  &                        \textbf{.71} &                            .07 &                       .41 &                            \textbf{.85} &                              .46 &                                  .47 &                             .17 &                                  .55 \\
& \textsc{I-L} & \textsc{Fix}   &                        .63 &                            .06 &                       .33 &                            .78 &                              .47 &                                  .54 &                             .19 &                                  .62 \\
& \textsc{Fix} & \textsc{I-L}   &                        .67 &                            .07 &                       .42 &                            .82 &                              .46 &                                  .60 &                             .22 &                                  .59 \\
& \textsc{I-L} & \textsc{I-L}    &                        .61 &                            .05 &                       .33 &                            .76 &                              .48 &                                  .65 &                             .24 &                                  .67 \\ \hline


\textsc{I-L} & \textsc{I-L} & \textsc{I-L} &                  .60 &                      .06 &                 .39 &                      .49 &                        \textbf{.57} &                            \textbf{.69} &                       \textbf{.41} &                            \textbf{.88} \\
\end{tabular}
\caption{Mean NormSuff and NormComp scores when we select at instance-level (I-L) a combination of the: (1) rationale length (\textsc{Len}); (2) feature scoring method (\textsc{Feat.}); and (3) rationale type (\textsc{Type}). \{\textsc{Type}\}-\textsc{Fix}-\textsc{Fix} and \{\textsc{Type}\}-\textsc{I-L}-\textsc{Fix} values are from the highest scoring feature scoring method (see Figure \ref{fig:fixed_len_var_feat}). \textbf{Bold} values denote the highest performing combination in column-wise (higher is better).} 
\label{tab:var_all}
\end{table}

\begin{figure*}[!t]
     \centering
     \begin{subfigure}[b]{0.25\textwidth}
         \centering
         \includegraphics[width=\textwidth, height=50mm]{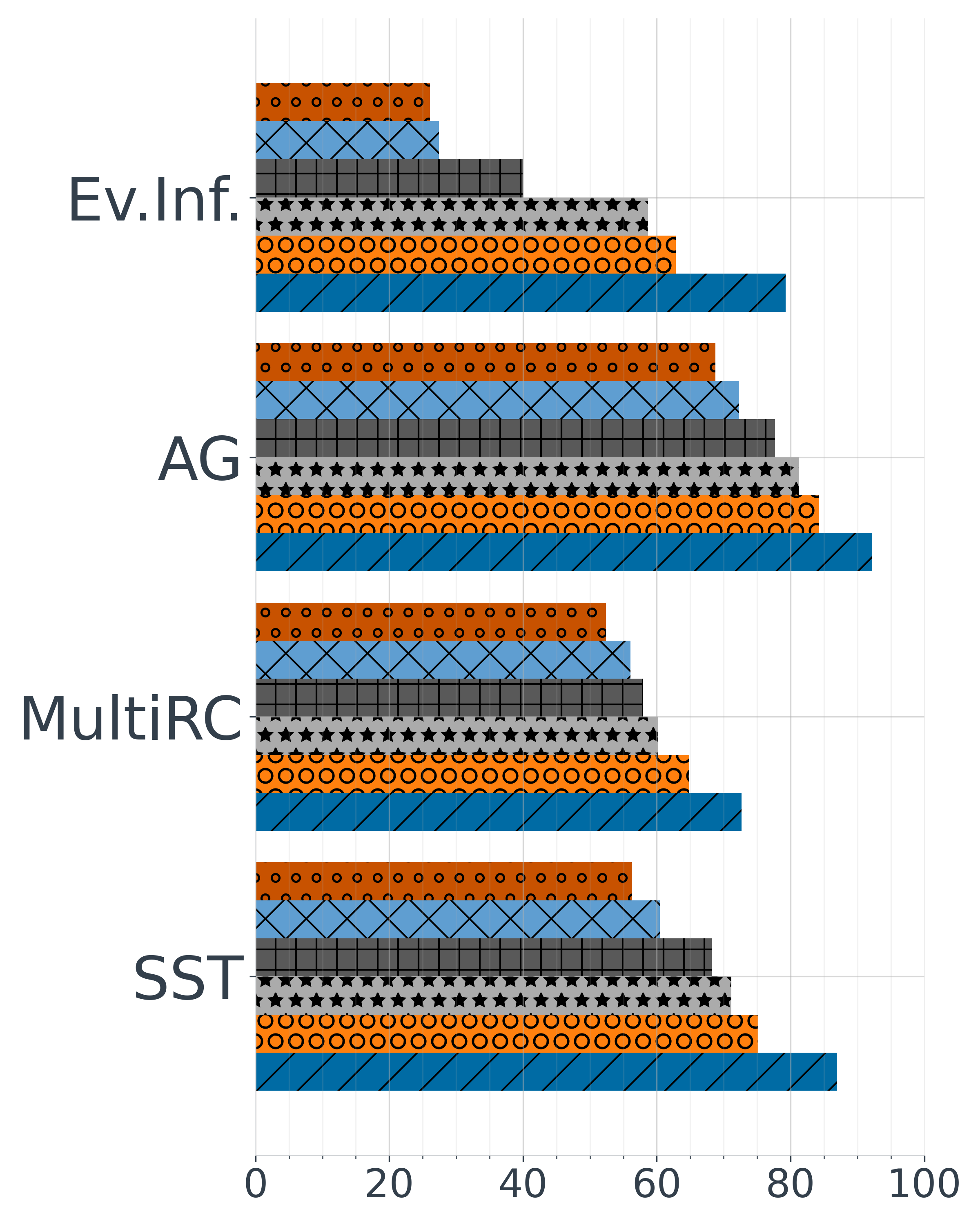}
         \caption{F1 macro}
     \end{subfigure}
     \hspace{0.0em}
     \begin{subfigure}[b]{0.25\textwidth}
         \centering
         \includegraphics[width=\textwidth, height=50mm]{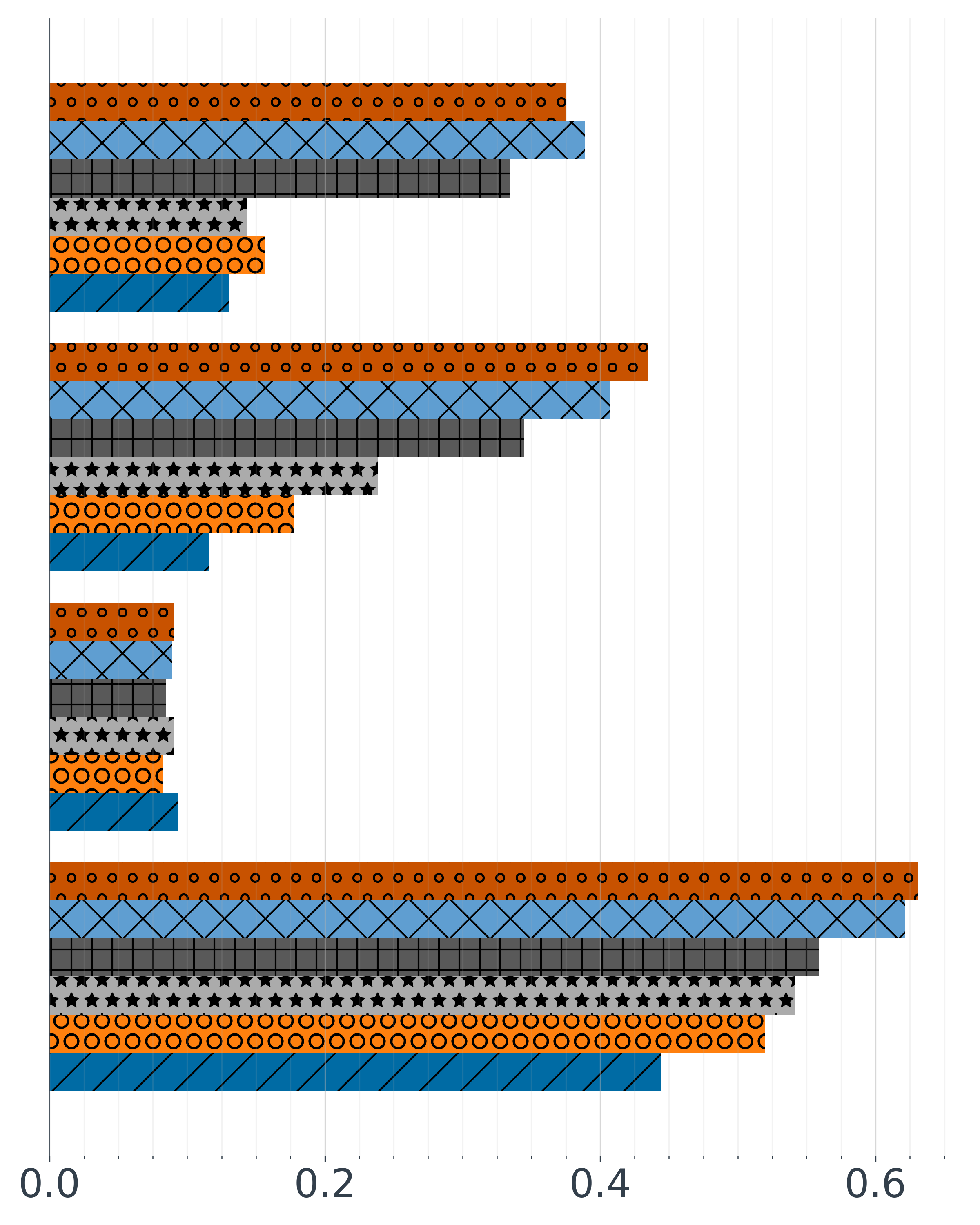}
         \caption{NormSuff}
     \end{subfigure}
     \hspace{0.0em}
     \begin{subfigure}[b]{0.25\textwidth}
         \centering
         \includegraphics[width=\textwidth, height=50mm]{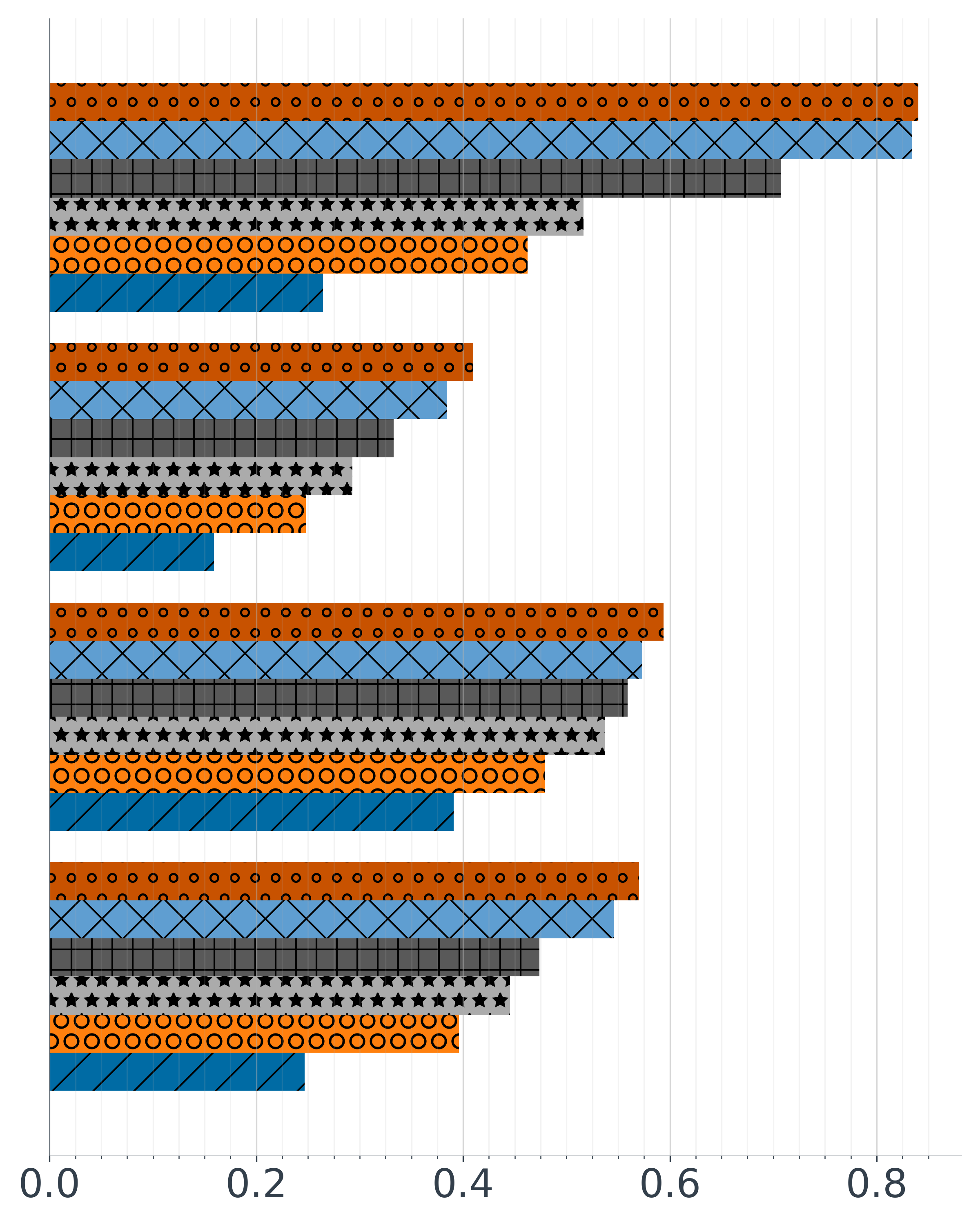}
         \caption{NormComp}
     \end{subfigure}
     \hspace{0.0em}
     \begin{subfigure}[b]{0.19\textwidth}
         \centering
        \raisebox{11mm}{
         \includegraphics[width=\textwidth]{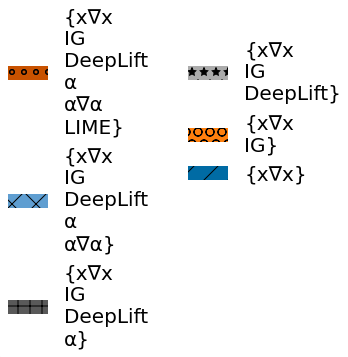}
        }
     \end{subfigure}
        \caption{F1 macro (lower is better), mean NormSuff and mean NormComp (higher is better), when extracting rationales with our approach given decreasing numbers of feature scoring methods.} 
        
        \label{fig:different_fixed_avid}
    
\end{figure*}

\subsection{Selecting Instance-specific Feature Scoring, Length and Type}

Table \ref{tab:var_all} shows mean NormSuff and NormComp scores when using our proposed method to select at instance-level (I-L) a combination of: (1) the feature scoring method (\textsc{Feat}); (2) the rationale length (\textsc{Len}); and (3) the rationale type (\textsc{Type}). For comparison, we also show scores of the best performing fixed (\textsc{Fix}) feature scoring function, rationale type and length (see Figure \ref{fig:fixed_len_var_feat}). For brevity we include results with F1 macro in App. \ref{appendix:results}, observing similar trends.

We first observe that the highest NormSuff scores across three datasets (SST, MRc, EvInf), are from the best performing fixed scoring method with fixed length and rationale type. Additionally, the best performing combination of our proposed approach for sufficiency is when we only select the feature scoring method keeping the length and type fixed. This combination results in the highest NormSuff scores in AG (.44 with \textsc{TopK} type compared to .42, which is the second best with \textsc{Contiguous}) and competitive NormSuff scores with the highest scoring combination (e.g. .82 in Ev.Inf. and \textsc{Contiguous} compared to .85). We assume that using combinations which include instance-specific lengths do not perform as well for sufficiency due to the shorter rationale length, which we have previously shown to partially degrade rationale sufficiency. 


Finally, our results demonstrate that we obtain highly comprehensive rationales when selecting at instance level all parameters (\textsc{Feat.} +  \textsc{Len} +  \textsc{Type}) using our approach. In fact, this results in higher NormComp scores compared to any other setting combination across all datasets. For example in M.Rc., selecting all parameters results in a NormComp score of .69 which is .22 units higher than the rationales extracted with fixed feature scoring method and length and type. This highlights the efficacy of our approach in extracting highly comprehensive rationales, without requiring strong a priori assumptions about rationale parameters.

\input{table_of_examples}


\subsection{Ablation Study
\label{sec:ablation}}

We finally perform an ablation study to examine the behavior and effectiveness of our approach by sequentially removing one feature scoring method at a time to measure changes in F1 macro, NormSuff and NormComp.
The intuition is that we should observe drops in faithfulness scores when removing feature attribution methods for our approach to be effective (i.e. we should extract more faithful rationales when having more feature scoring options to choose from). Figure \ref{fig:different_fixed_avid} shows the results. 

We first observe that removing one feature scoring method at a time results in increases in F1 macro (lower is better) and drops in NormComp scores (higher is better). This demonstrate that the faithfulness of the rationales extracted with our approach deteriorates as the number of feature scoring methods becomes smaller highlighting the efficacy of our proposed approach. 
For example, in Ev.Inf. by removing $\alpha\nabla\alpha$ results in a drop of .14 in mean NormComp (.84 when including $\alpha\nabla\alpha$ compared to .70 without it). 
On the other hand, we also observe that our method can still benefit from feature scoring methods that achieve low NormComp scores when used standalone, resulting in improvements in comprehensiveness and drops in F1 macro (e.g. $\alpha$ in SST).
This indicates that our approach steadily improves rationale faithfulness for model's predictions given a larger pool of available feature scoring methods.

Results show a deterioration in NormSuff scores as the number of feature scoring methods becomes smaller, showing that our method results in more sufficient rationales when presented with a larger list of available feature scoring methods in the majority of the datasets. 
We hypothesize that this is not true for MultiRC due to the already low NormSuff scores of the rationales (e.g. no more than 0.12).
By using all six feature scoring methods, our approach produces highly sufficient rationales and is comparable to the set achieved the highest sufficiency. For example in Ev.Inf. using all feature scoring methods results to a NormSuff score of approximately .38 compared to the highest scoring feature scoring set (all except LIME) and the lowest scoring ($\mathbf{x}\nabla\mathbf{x}$) which achieved .39 and .15 respectively.

We also tested different combinations of feature scoring methods with similar observations (see App. \ref{appendix:results_feat_scoring_diff}). Finally, we experimented with doubling the upper bound of the rationale length (from $N$ to $2 \times N$) for both fixed length rationales and our proposed approach. Our approach still yielded more comprehensive rationales compared to the fixed-length ones that were also highly sufficient (results included in App. \ref{appendix:results_double_n}).

\section{Qualitative Analysis
\label{sec:qualitative}}

Table \ref{fig:annotations} shows examples of the qualitative comparison between our approach (Ours) for selecting at instance-level (I-L) a combination of the: (1) rationale length (\textsc{Len}); (2) feature scoring method (\textsc{Feat} against our baseline of fixed-length rationales from a fixed feature scoring method. 

\paragraph{Concise rationales:} Example 1 presents an instance from \textsc{AG}.
Our approach extracts a rationale that is six tokens shorter than the one with fixed length while also achieving a higher NormComp score. However, the fixed length rationale scores higher in NormSuff. We can assume from this that sufficiency positively correlates with rationale length.

\paragraph{Error analysis:} Our assumption is that if a model makes a wrong prediction, we should be able to extract the rationale that better demonstrates what led to a wrong prediction. Example 2 shows an instance from \textsc{Ev.Inf.}, where the model has wrongly predicted that ``Lenke scores at 12 months'' have `increased significantly' instead of the correct `no significant difference'. Surprisingly, both rationales recorded maximum scores (1.0) in NormSuff and NormComp.
We observe that the correct answer is included in the fixed length rationale, however the model made a wrong prediction. 
 On the contrary, our rationale highlights something directly related to its prediction. 

Example 3 presents an instance from SST, where the fixed-length rationale and the instance-specific rationale (ours) attend at different sections of the text. Our rationale scored lower for NormSuff, however we observe that it aligns more closely with the predicted sentiment. 

\paragraph{When using a fixed pre-defined length is not sufficient:} Example 4 presents a different scenario, where the fixed-length rationale for \textsc{SST} is at 20\% whilst the upper bound $N$ for our rationale is at 40\%. The intuition is that in certain cases a fixed rationale length might not be sufficient for all instances to explain a prediction. We argue that our approach highlighted something more informative for the task (``incredibly dull'' compared to  ``incredibly''), due to removing the restriction of a pre-defined fixed length.

\section{Conclusions}

We have proposed a simple yet effective approach for selecting at instance-level (1) feature scoring method; (2) length; and (3) type of the rationale. We empirically demonstrated that rationales extracted with our approach are significantly more comprehensive and highly sufficient, while being shorter compared to rationales extracted with a fixed feature scoring method, length and type.
Finally, we consider our work an important step towards instance-level faithful rationalization while finding the most sufficient rationale, an interesting direction for future work.

\section*{Acknowledgments}
NA is supported by EPSRC grant EP/V055712/1, part of the European Commission CHIST-ERA programme, call 2019 XAI: Explainable Machine Learning-based Artificial Intelligence.


\appendix

\bibliography{aaai22}

\clearpage
\newpage

\section{Model Hyperparameters
\label{appendix:model_hyperparameters}}

Table \ref{tab:model_hyperparameters} presents the hyper-parameters used to train the models across different datasets, along with F1 macro performance on the development set. Models where finetuned across 3 runs for 5 epochs. We implement our models using the Huggingface library \citep{Wolf2019HuggingFacesTS} and use default parameters of the \textsc{AdamW} optimizer apart from the learning rates. We use a linear scheduler with 10\% of the steps in the first epoch as warmup steps. Experiments are run on a single Nvidia Tesla V100 GPU.

\renewcommand*{\arraystretch}{1}
\begin{table}[!b]
\setlength\tabcolsep{2.5pt}
\small
\centering
\begin{tabular}{l||ccc|c}
\textbf{Dataset}   & Model & $lr^m$ & $lr^c$ &  F1 \\ \hline
SST                & bert-base & 1e-5 & 1e-4     &  90.7  $\pm$ 0.2   \\
AG         & bert-base & 1e-5 & 1e-4     &  93.3 $\pm$ 0.0      \\
Ev.Inf.             & scibert & 1e-5 & 1e-4   & 82.5 $\pm$ 0.9       \\
M.RC           & roberta-base & 1e-5 & 1e-4    &  76.3 $\pm$ 0.2      \\
\end{tabular}
\caption{Model and their hyper-parameters for each dataset, including learning rate for the model ($lr^m$) and the classifier layer ($lr^c$)  and F1 macro scores on the development set across three runs.}
\label{tab:model_hyperparameters}
\end{table}

\section{Divergence metrics ($\boldsymbol \Delta$)}
\label{appendix:divergence_metrics}

To compute the value $\delta$ for how much $\mathcal{Y}^m$ differs from $\mathcal{Y}$, we consider four divergence measures ($\Delta$), also previously used in literature \citep{robnik2008explaining, jain2019attention, wiegreffe2019attention}:

\paragraph{Kullback Leibler (KL)}: A non-symmetric divergence measure of how a particular distribution divergences from a reference distribution:

\begin{equation}
    KL(\mathcal{Y}||\mathcal{Y^*}) = \mathcal{Y} (\log(\mathcal{Y} - \log(\mathcal{Y}^m))
\end{equation}

\paragraph{Jensen-Shannon (JSD)}: A symmetric divergence metric based on the KL divergence of two distributions from their mean:

\begin{equation}
    JSD(\mathcal{Y}||\mathcal{Y}^m) = \frac{1}{2}(KL(\mathcal{Y}||\mathbf{\mu}) + \frac{1}{2}(KL(\mathcal{Y}^m||\mu))
\end{equation}

\noindent where $\mu$ is the average distribution of $\mathcal{Y}$ and $\mathcal{Y}^m$. 

\paragraph{Perplexity (\textsc{Perp.})}: A measure of how well a model can predict a sample, where:

\begin{equation}
    PERP(\mathcal{Y}||\mathcal{Y}^m) = \exp^{\mathcal{H(\mathcal{Y} , \mathcal{Y}}^m)}
\end{equation}

\noindent where we consider $\mathcal{Y}$ as the ground truth and $H(\mathcal{Y}||\mathcal{Y}^m)$ is the Cross Entropy Loss.

\paragraph{Class Difference (\textsc{ClassDiff})}: The direct difference between the predicted class probability from the model with full text ($\mathbf{x}$) and the same class probability with reduced text ($\mathbf{x}_{\setminus\mathcal{R}}$) :

\begin{equation}
    CLASSDIFF(\mathcal{Y}||\mathcal{Y}^m) = p(\hat{y}| \mathbf{x})- p (\hat{y}|\mathbf{x}_{\backslash\mathcal{R}})
\end{equation}

\noindent where $\hat{y} = \text{arg max}(\mathcal{Y})$.

\section{Computed Instance-Specific Lengths
\label{appx:computed_variable_lengths}}

\renewcommand*{\arraystretch}{1}
\begin{table}[!t]
\small
\centering
\setlength{\tabcolsep}{2pt}
\begin{tabular}{cl||cccccc|c}
\% & {} &   $\mathbf{x}\nabla\mathbf{x}$ &    IG & DeepLift &  LIME &     $\alpha$ &   $\alpha\nabla\alpha$ &  Avg. \\ \hline \hline
\parbox[t]{3mm}{\multirow{4}{*}{\rotatebox[origin=c]{90}{\textsc{TopK}}}} & SST          &  15.8 &  16.3 &     15.7 &  16.9 &  15.5 &  16.6 &  16.1 \\
& AG          &  14.5 &  16.5 &     15.2 &  16.3 &  16.3 &  16.3 &  15.8 \\
& Ev.Inf.       &   7.7 &   7.7 &      8.4 &   7.4 &   6.6 &   7.0 &   7.5 \\
& MultiRC       &  14.1 &  14.6 &     16.1 &  13.4 &  15.9 &  16.0 &  15.0 \\ \hline
\parbox[t]{3mm}{\multirow{4}{*}{\rotatebox[origin=c]{90}{\textsc{Cont.}}}} & SST     &  15.5 &  15.6 &     15.4 &  15.9 &  14.7 &  15.7 &  15.5 \\
& AG      &  14.0 &  15.7 &     14.9 &  15.0 &  14.5 &  15.1 &  14.9 \\
& Ev.Inf. &   7.2 &   6.9 &      7.6 &   7.3 &   6.6 &   6.9 &   7.1 \\
& MultiRC &  14.0 &  15.1 &     15.9 &  13.6 &  16.1 &  16.1 &  15.1 \\
\end{tabular}
\caption{Average instance-specific rationale lengths (as a percentage \%) computed using JSD (as $\Delta$), across instances for \textsc{TopK} and \textsc{Contiguous} rationale types.}
\label{tab:rationale_ratios}
\end{table}

\begin{table*}[!t]
    \centering
    \setlength{\tabcolsep}{3pt}
    \begin{tabular}{ll||cccc|cccc|cccc}
        {} & {} & \multicolumn{4}{c|}{@Token} & \multicolumn{4}{c|}{@2\%} & \multicolumn{4}{c}{@5\%} \\    
    {} & {} &  F1  &  Suff.  &  Comp. &  Rat-Len.  & F1  &  Suff.  &  Comp. &  Rat-Len.  & F1  &  Suff.  &  Comp. &  Rat-Len. \\ \hline \hline
     \parbox[t]{3mm}{\multirow{4}{*}{\rotatebox[origin=c]{90}{\textsc{TopK}}}}  & SST           &                   57.22 &                       0.59 &                             0.55 &                          17.15 &           57.22 &               0.59 &                     0.55 &                  17.15 &           57.34 &               0.59 &                     0.55 &                  17.23 \\
     
    & AG            &                   69.90 &                       0.37 &                             0.39 &                          17.05 &           69.93 &               0.37 &                     0.39 &                  17.06 &           70.12 &               0.38 &                     0.39 &                  17.52 \\
    
    & Ev.Inf.       &                   21.09 &                       0.35 &                             0.87 &                           6.52 &           21.74 &               0.36 &                     0.86 &                   7.25 &           23.22 &               0.37 &                     0.84 &                   8.28 \\
    
    &MultiRC       &                   47.63 &                       0.07 &                             0.63 &                          13.59 &           48.76 &               0.07 &                     0.62 &                  14.31 &           49.92 &               0.08 &                     0.61 &                  15.37 \\ \hline
    
    \parbox[t]{3mm}{\multirow{4}{*}{\rotatebox[origin=c]{90}{\textsc{Cont.}}}}  & SST     &                   66.43 &                       0.61 &                             0.48 &                          16.01 &           66.43 &               0.61 &                     0.48 &                  16.01 &           66.48 &               0.61 &                     0.48 &                  16.21 \\
    
    &AG      &                   85.27 &                       0.33 &                             0.24 &                          15.04 &           85.27 &               0.33 &                     0.24 &                  15.06 &           85.42 &               0.34 &                     0.24 &                  16.00 \\
    
    &Ev.Inf. &                   37.03 &                       0.68 &                             0.70 &                           6.12 &           40.66 &               0.76 &                     0.67 &                   7.08 &           46.20 &               0.84 &                     0.62 &                   8.37 \\
    
    &MultiRC &                   43.38 &                       0.05 &                             0.66 &                          12.29 &           45.12 &               0.05 &                     0.65 &                  13.22 &           47.87 &               0.06 &                     0.63 &                  14.42 \\
\end{tabular}
\caption{F1 macro (lower is better), NormSuff (higher is better) and NormComp (higher is better) for our rationales with instance-specific length and feature scoring method at each instance, when computing $\delta$ at: (1) each token (@Token); (2) at every 2\% (@ 2\%) and at every 5\% (@ 5\%). We also include average rationale lengths for helping with the analysis.}
\label{tab:faith_at_different_@}
\end{table*}

In Table \ref{tab:rationale_ratios} we present the average computed rationale lengths using JSD across each feature scoring method, for each dataset and rationale type. To extract rationales we use an upper bound $N$, our pre-defined fixed rationale length (as indicated in Table \ref{tab:data_characteristics} and as defined by \citep{jain2020learning}), to make it comparable with our baseline. 

We first observe that rationales extracted with our proposed approach are on average shorter than the a priori set rationale ratio. In datasets such as \textsc{SST} and \textsc{AG} where we have short sequences on average (See Table \ref{tab:data_characteristics}), these differences are negligible, as a 4\% decrease translates to approximately having a single token less in the rationale. This strengthens our initial hypothesis that certain instances might not require as many tokens to successfully explain a prediction.

In datasets with longer sequences lengths on average (\textsc{Ev.Inf.} and \textsc{M.RC}) such differences are more evident. For example with LIME and \textsc{Contiguous} rationales in \textsc{M.RC}, our proposed approach results in rationales which are on average 6\% shorter than the fixed length rationales. This translates to approximately 20 less tokens to form a rationale. What is particularly interesting is that by observing the standard deviations for \textsc{M.RC} and \textsc{Ev.Inf.}, we notice that the vast majority of instances does not exhaust the upper bound $N$ to form a rationale. We consider this particularly important for longer sequences, as often when acquiring an explanation for a model's prediction it is desirable to avoid a noisy interpretation of why a model predicted a particular class.

\section{Alternative Formulation of Instance-Specific Length Rationales
\label{appendix:alternative_delta}}

We also examined introducing a thresholding approach in early experimentation when extracting instance-specific length rationales, whereby $\delta_{prev} - \delta_{max} < thresh$, similar to early stopping to avoid exhausting the upper bound $N$. Early results suggested that for datasets with shorter length sequences (\textsc{SST} and \textsc{AG}) this approach performed worse albeit comparably. In datasets with longer sequence lengths (\textsc{Ev.Inf.} and \textsc{M.Rc}), the threshold approach performed poorly. On a closer inspection this was attributed to finding a threshold too early in the sequence thus not capturing all the necessary information. We experimented with \textsc{JSD} and the following thresholds : \{1e-4, 1e-3, 1e-2\}.

We considered using patience to avoid such naive thresholding, however the computation time would increase significantly. The reason being that we conduct experiments at instance-level, in batches. Introducing patience, would entail that we conduct it using a single instance at a time thus increasing computations by the batch size. As such we have not conducted this approach as it would be computationally expensive.

\begin{table}[!b]
    \centering
    \setlength{\tabcolsep}{5pt}
    \begin{tabular}{llc||cc|cc}
    {} & {} & @Token & \multicolumn{2}{c|}{@2\%} & \multicolumn{2}{c}{@5\%} \\
    {} & {} &  (s) &  (s) &  R.I. & (s) & R.I.  \\
    \hline \hline
    \parbox[t]{3mm}{\multirow{4}{*}{\rotatebox[origin=c]{90}{\textsc{TopK}}}} &SST           &             0.05 &     0.05 &        1.0 &     0.05 &        1.0 \\
    &AG            &             0.29 &     0.29 &        1.0 &     0.15 &        1.9 \\
    &Ev.Inf.       &             1.99 &     0.26 &        7.7 &     0.11 &       18.1 \\
    &MultiRC       &             3.07 &     0.51 &        6.0 &     0.21 &       14.6 \\ \hline
\parbox[t]{3mm}{\multirow{4}{*}{\rotatebox[origin=c]{90}{\textsc{Cont.}}}} & SST     &             0.06 &     0.06 &        1.0 &     0.06 &        1.0 \\
    &AG      &             0.37 &     0.38 &        1.0 &     0.19 &        1.9 \\
    &Ev.Inf. &             2.59 &     0.33 &        7.8 &     0.13 &       19.9 \\
    & MultiRC &             3.72 &     0.65 &        5.7 &     0.26 &       14.3 \\
    \end{tabular}
    \caption{Average time taken (s) to extract a rationales of instance-specific length per instance, when computing $\delta$ at: (1) each token (@Token); (2) at every 2\% (@ 2\%) and at every 5\% (@ 5\%), where lower time is better. We also denote relative improvements (R.I.) where higher is better.}
    \label{tab:complexity_reduction}
\end{table}

\section{Reducing Time Complexities
\label{appendix:complexities}}

Albeit significantly faster to computing LIME scores, selecting a rationale length at each instance in a dataset can be computationally expensive when we compute $\delta$ for every token, being similar to counting decision flips \citep{nguyen-2018-comparing, serrano2019attention, atanasova2020diagnostic}. This takes into consideration that we have to perform a forward pass for every token until we reach $N$ tokens, for each feature attribution approach $\Omega$. In the following segments we describe approaches to reduce computational times.

\paragraph{Reducing search granularity: } Similar to \citet{atanasova2020diagnostic}, we can reduce significantly computation times by reducing the granularity of our search. In our implementation we describe masking each token or n-gram sequentially, which can be altered to skip tokens. For example, consider a sequence with 200 tokens and an upper-bound $N=$ 20\% and as such $N_t = 40$. Instead of computing $\delta$ for each token, we can compute it for every 5 tokens and as such reducing complexity by 5. Similarly we can compute $\delta$ at every 2\% of the sequence until we reach $N$. For example for \textsc{Ev.Inf.}, where $N=$10\% we compute $\delta$ at every \{2\%, 4\%, $\hdots$\, 10\%\} thus keeping the forward passes constant across instances.

\begin{figure*}[!t]
     \centering
     \begin{subfigure}[b]{0.29\textwidth}
         \centering
         \includegraphics[width=\textwidth]{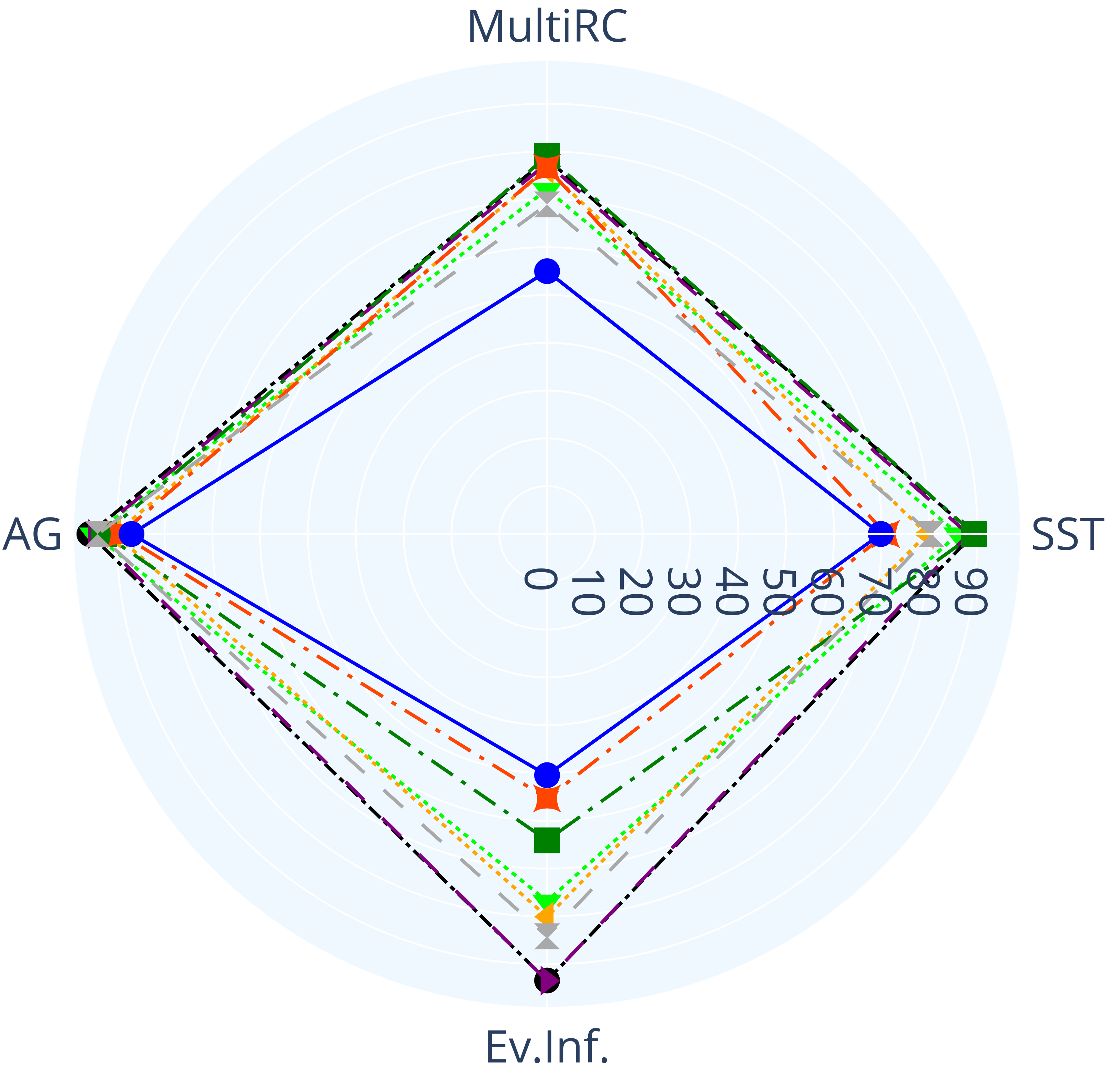}
         \caption{F1 macro}
     \end{subfigure}
     \hspace{0.0em}
     \begin{subfigure}[b]{0.29\textwidth}
         \centering
         \includegraphics[width=\textwidth]{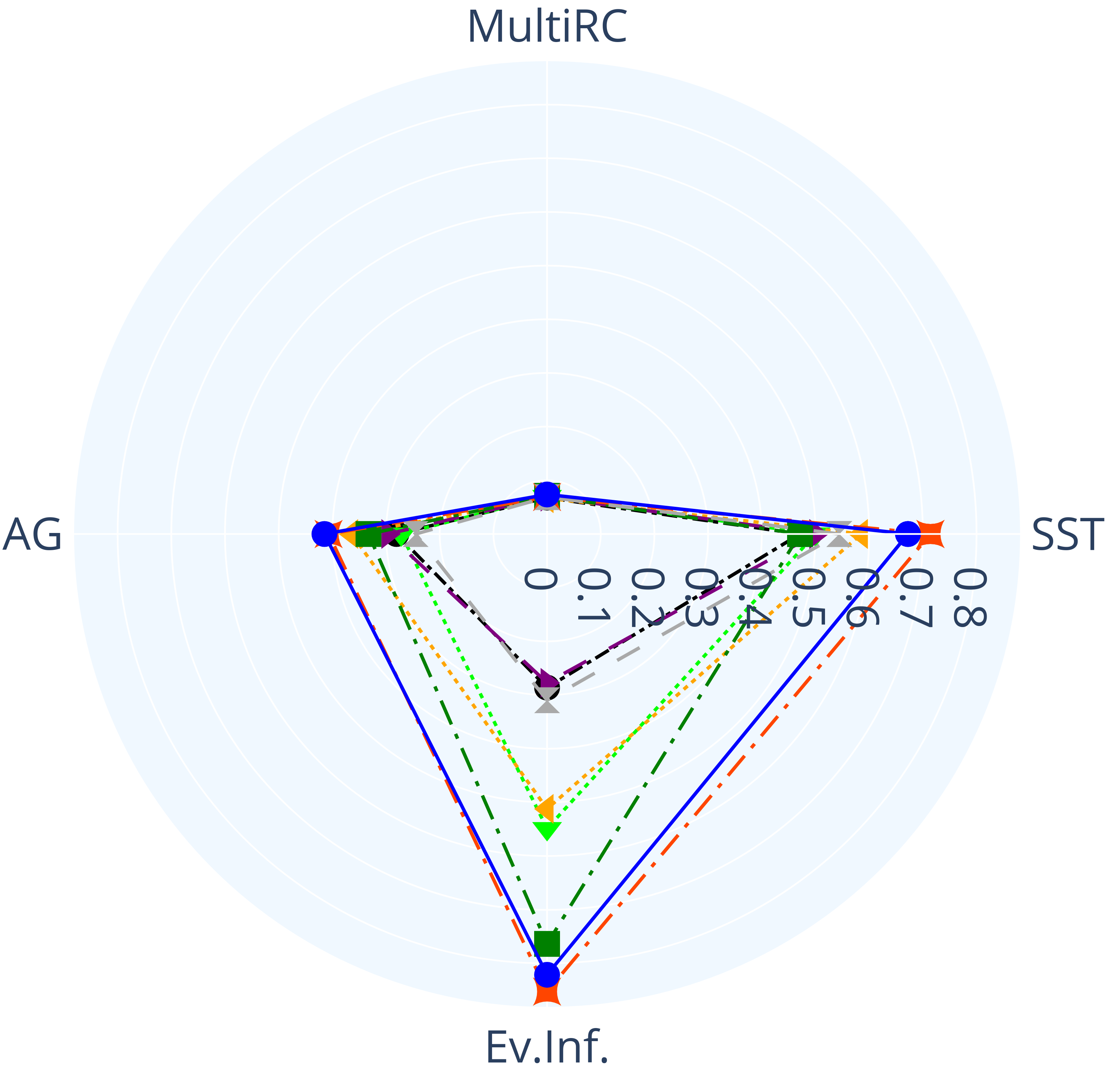}
         \caption{Sufficiency}
     \end{subfigure}
     \hspace{0.0em}
     \begin{subfigure}[b]{0.29\textwidth}
         \centering
         \includegraphics[width=\textwidth]{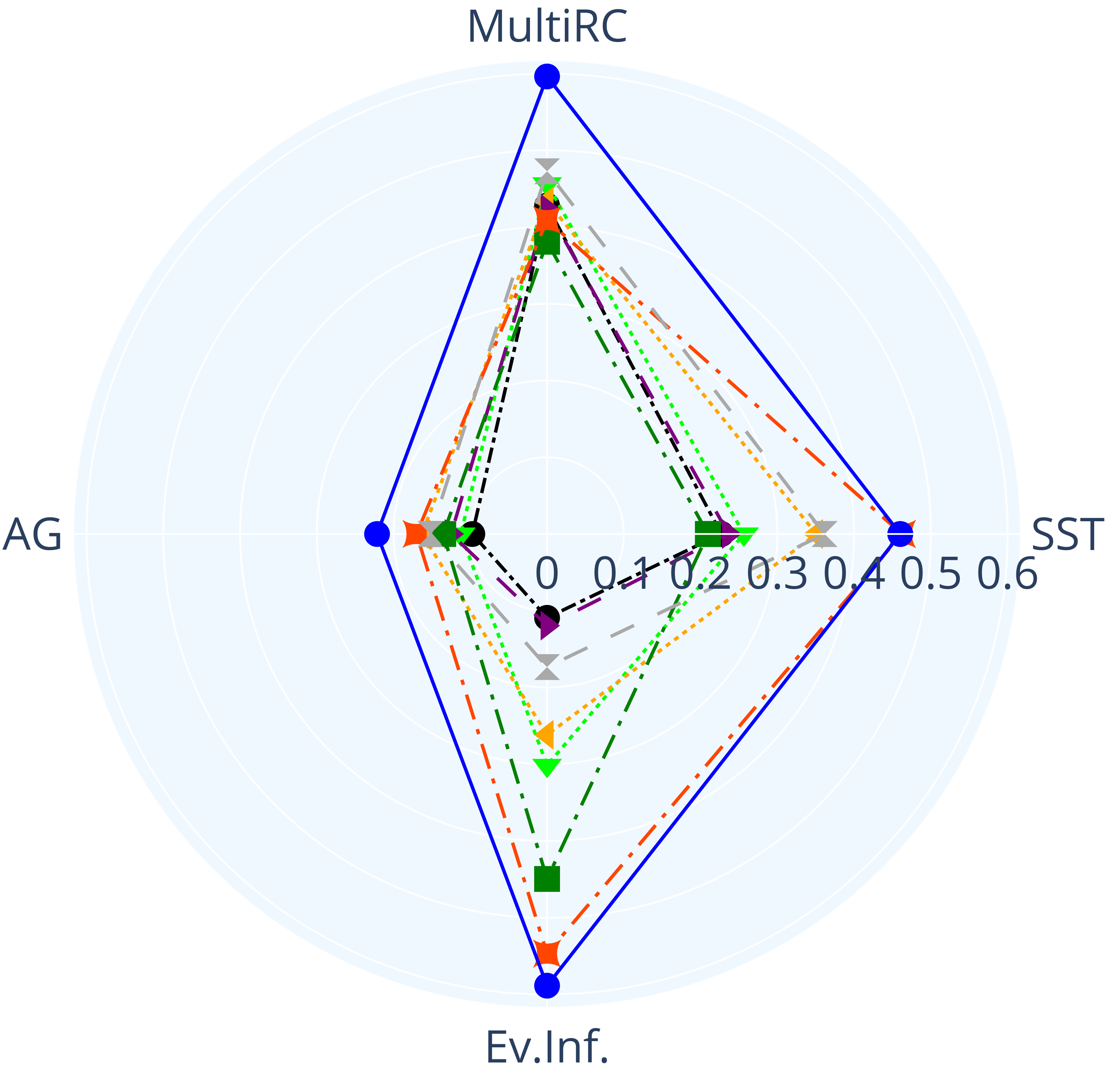}
         \caption{Comprehensiveness}
     \end{subfigure}
     \hspace{0.0em}
     \begin{subfigure}[b]{0.10\textwidth}
         \centering
         \includegraphics[width=\textwidth]{new_images/selecting_feat_radar/legend-var-feat-score.png}
     \end{subfigure}
        \caption{F1 macro (lower is better), mean NormSuff (higher is better) and mean NormComp (higher is better) when using any single feature scoring method across all instances in a dataset and our proposed method of selecting a feature scoring method for each instance (OURS).}
        
        \label{fig:fixed_len_var_feat_cont}
    
\end{figure*}
In Table \ref{tab:complexity_reduction}, we present average time taken (in seconds (s)) to extract a rationale of instance-specific length for each instance,when computing $\delta$ at: (1) each token (@Token); (2) at every 2\% (@ 2\%) and at every 5\% (@ 5\%) using JSD. We observe that in datasets with a short average length of instance when moving from @Token to @2\% does not reduce time, but results in significant reductions in computations with MultiRc ($\sim$6x R.I.) and Evinf ($\sim$8x R.I.). As expected, these are further reduced when reducing the granularity to searching @5\%, with AG recording a $\sim$2x R.I., MultiRC $\sim$14.5 R.I. and Ev.Inf. $\sim$19\% R.I.

In Table \ref{tab:faith_at_different_@} we present the faithfulness performance as we reduce the granularity of our search. Results suggest that by reducing granularity, NormComp scores reduce whilst F1 macro performances increase, suggesting a reduction in faithfulness. However, we observe that moving from @Token to @2\% this reduction is negligible considering the significantly improved computational times. However, as expected moving from @Token to @5\% performance degrades rapidly and as such 2\% seems like a more appropriate step to consider. Unsurprisingly, with increasing step-size we observe increases in the computed rationale lengths, which leads to an increase in sufficiency.

\paragraph{Combining feature scoring rankings:} We considered further reducing our computation time by merging importance scores from all feature scoring methods. The intuition is that we obtain a combined ranking and avoid selecting the best feature scoring method at each instance and computing a rationale length for all feature scoring methods. We attempted this by averaging the normalized importance scores for each sequence from all the feature scoring methods, however as expected results where not comparable (63.2 average F1 macro compared to 54.4) with our proposed approach or even our best performing baseline. 

\section{Comparing divergence metrics
\label{appendix:comparing_divmetrics}}

We first compare the effectiveness of divergence metrics in computing a rationale with instance-specific length and selecting the best feature scoring method at instance level. Table \ref{tab:full_stack_all} presents F1 macro (lower is better), NormSuff and NormComp (higher is better) macro scores for our proposed instance-specific length rationales from the best feature scoring method at instance level. 

Results demonstrate that all divergence metrics perform comparably with the exception of perplexity. The remainder of the metrics result in similar scores for NormSuff, NormComp and F1 macro, with JSD and ClassDiff having a slight edge over KLDIV.

\begin{table}[!t]
    \centering
    \begin{tabular}{ll||ccc}
        {} & $\Delta$ &  NormSuff &  NormComp &  F1 \\ \hline \hline
        \parbox[t]{3mm}{\multirow{4}{*}{\rotatebox[origin=c]{90}{\textsc{TopK}}}} & JSD        &         0.38 &               0.64 &                        46.03 \\
        & KLDIV      &         0.38 &               0.63 &                        46.40 \\
        & ClassDiff  &         0.38 &               0.63 &                        45.02 \\
        & Perp. &         0.35 &               0.55 &                        52.22 \\ \hline
        \parbox[t]{3mm}{\multirow{4}{*}{\rotatebox[origin=c]{90}{\textsc{TopK}}}} &  JSD        &         0.38 &               0.64 &                        46.03 \\
        & KLDIV    &         0.38 &               0.63 &                        46.40 \\
        & ClassDiff  &         0.38 &               0.64 &                        45.02 \\
        & Perp. &         0.35 &               0.55 &                        52.22 \\
        \end{tabular}
    \caption{NormSuff (higher is better), NormComp (higher is better) and F1 macro (lower is better) when using different divergence metrics to select the rationale length and feature scoring method($\Delta$)}
    \label{tab:full_stack_all}
\end{table}

\begin{figure*}[!t]
     \centering
     \begin{subfigure}[b]{0.25\textwidth}
         \centering
         \includegraphics[width=\textwidth]{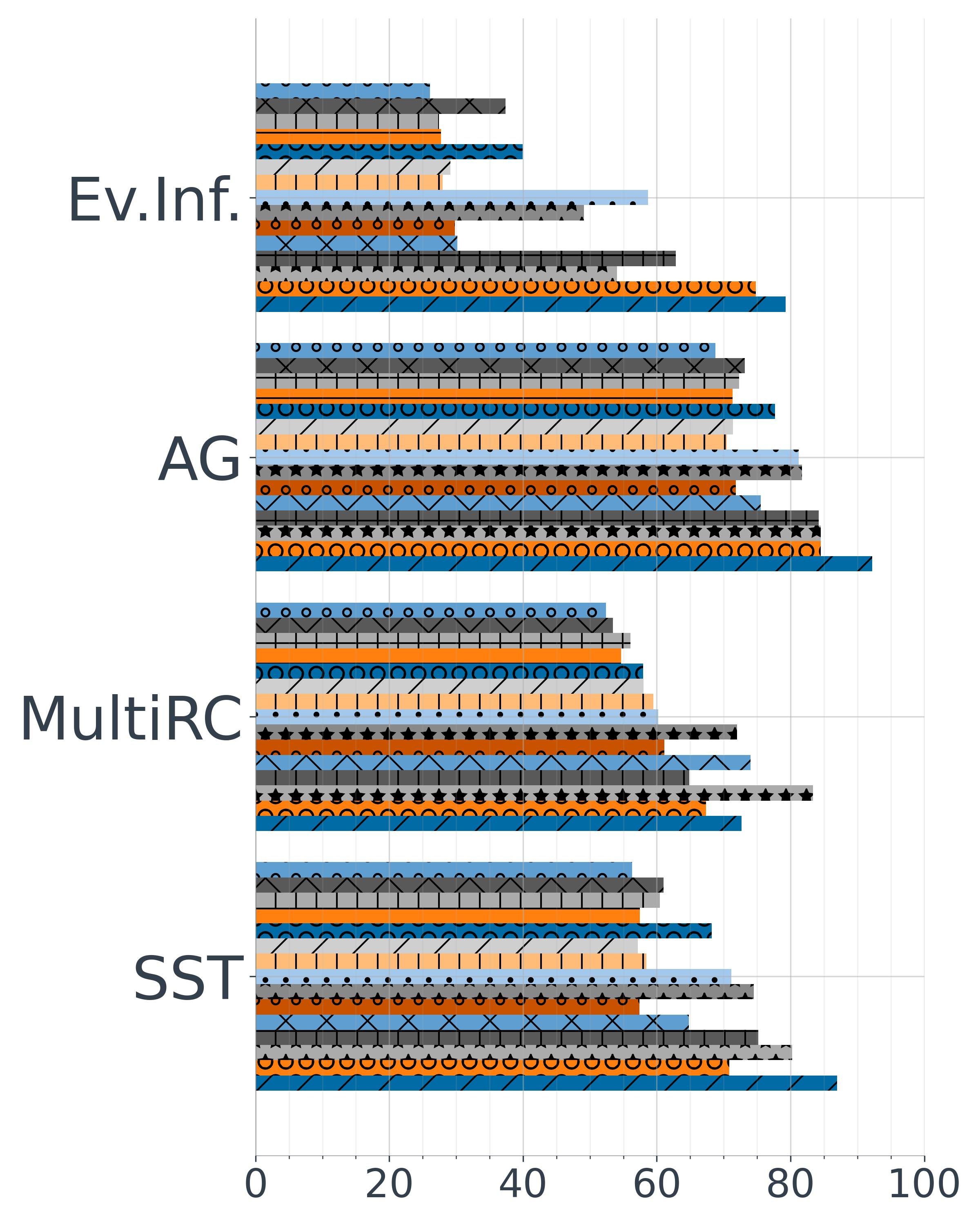}
         \caption{F1 macro}
     \end{subfigure}
     \hspace{0.0em}
     \begin{subfigure}[b]{0.25\textwidth}
         \centering
         \includegraphics[width=\textwidth]{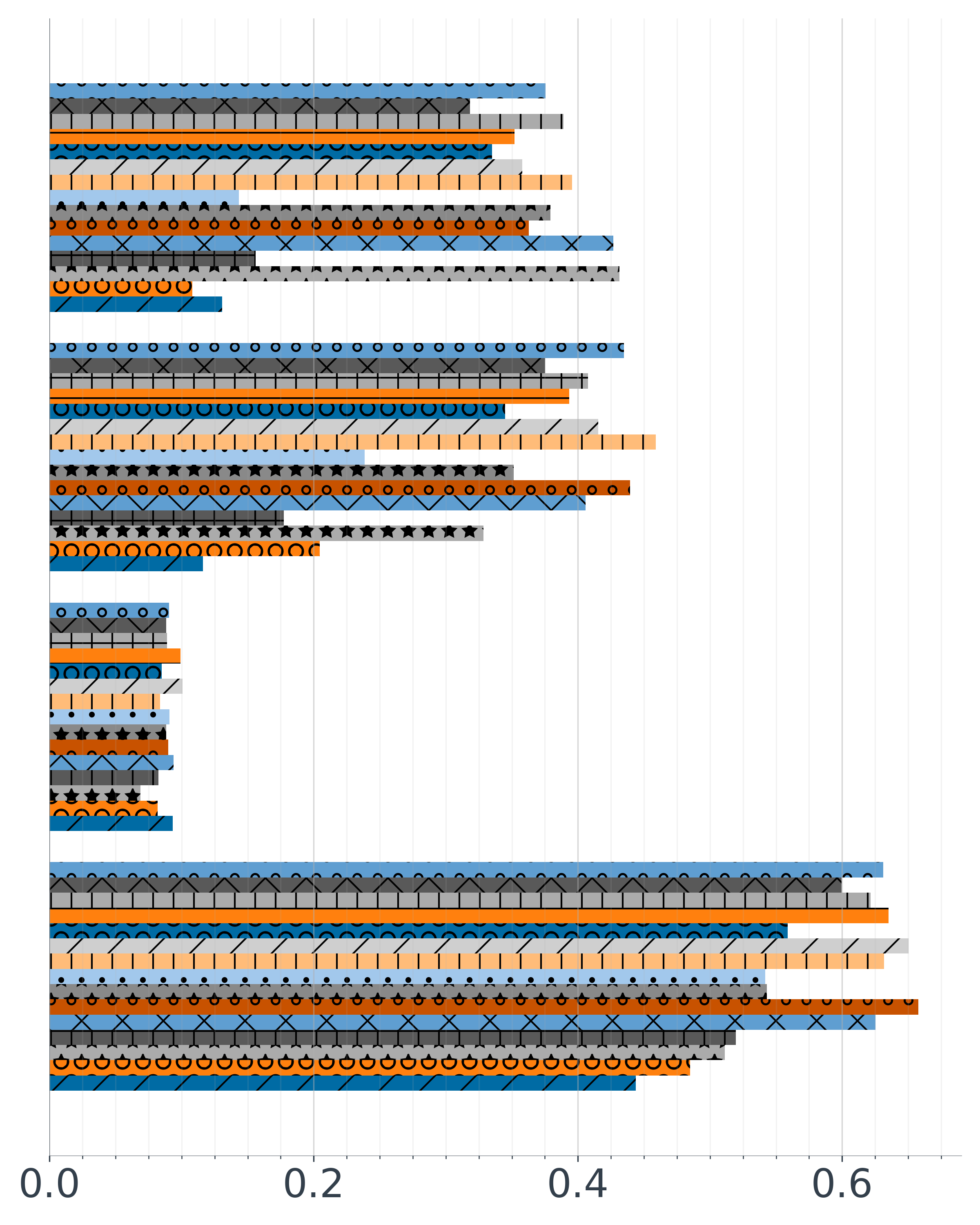}
         \caption{NormSuff}
     \end{subfigure}
     \hspace{0.0em}
     \begin{subfigure}[b]{0.25\textwidth}
         \centering
         \includegraphics[width=\textwidth]{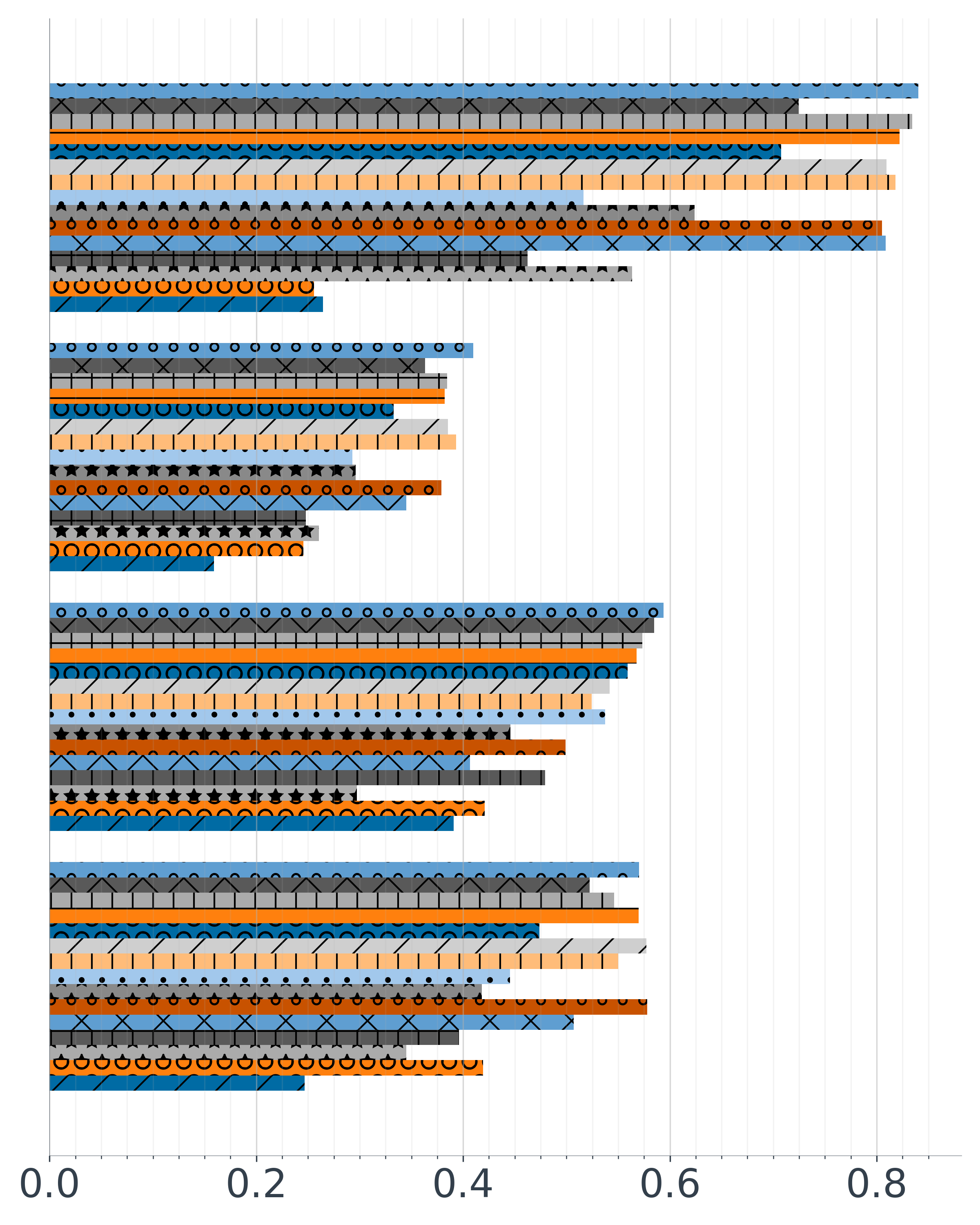}
         \caption{NormComp}
     \end{subfigure}
     \hspace{0.0em}
     \begin{subfigure}[b]{0.18\textwidth}
         \centering
        \raisebox{11mm}{
         \includegraphics[width=\textwidth]{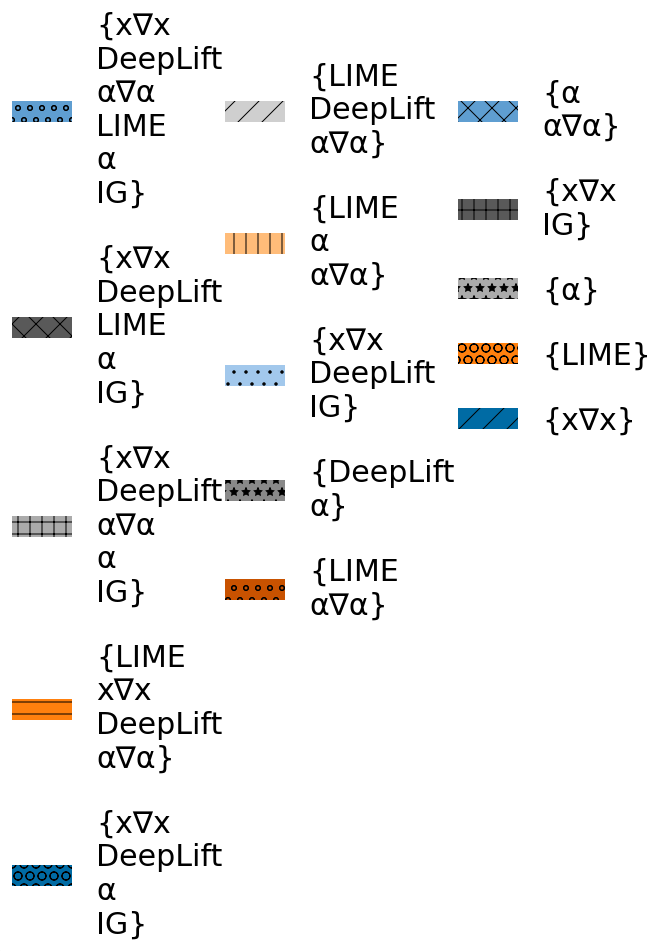}
        }
     \end{subfigure}
        \caption{F1 macro (lower is better), mean NormSuff and mean NormComp (higher is better), when extracting rationales with our approach given decreasing numbers of feature scoring methods.}
        
        \label{fig:different_fixed_avid_appendix}
    
\end{figure*}

\section{Additional Results
\label{appendix:results}}

\subsection{Selecting a Feature Scoring Method
\label{appendix:results_feat_scoring_cont}} 

Figure \ref{fig:fixed_len_var_feat_cont} shows F1 macro (lower is better), mean NormSuff and mean NormComp (higher is better), when using a single feature scoring method globally (across all instances in a dataset) to extract rationales and when we use our proposed approach to select a feature scoring method for each instance (OURS) for contiguous type rationales. Overall, results are similar to those observed using \textsc{TopK}, with our approach resulting to rationales with significantly higher NormComp scores and lower F1 macro scores, whilst still retaining high sufficiency as demonstrated by the high NormSuff scores. Overall, with \textsc{Contiguous} rationales our approach results in slight improvements in NormSuff scores over \textsc{TopK} when using our proposed method. 

\subsection{Different Sets of Feature Scoring Methods
\label{appendix:results_feat_scoring_diff}}

Figure \ref{fig:different_fixed_avid_appendix} shows F1 macro (lower is better), mean NormSuff and mean NormComp (higher is better) scores, when using our proposed method to select from different sets of feature scoring methods. 

Overall, we observe similar results to the analysis in \S\ref{sec:ablation} with our approach reaching peak performance in NormComp (highest) and F1 macro (lowest) with the set including all the feature scoring methods. Also, similar to the analysis in the main body, we observe that again our approach results in highly sufficient rationales when presented with the largest set of feature scoring methods. 

What is demonstrated from this more detailed analysis, is that certain feature scoring methods contribute more to our approach compared to others. For example LIME and $\alpha\nabla\alpha$ in Ev.Inf. is enough to reach a NormComp score of .80 compared to the best of .84, so only .04 points behind. We find a correlation between the feature scoring methods when they are performing well globally (see Fig \ref{fig:fixed_len_var_feat}) and when they are improving drastically our proposed approach, as they are essentially giving a good head-start to our approach.

\subsection{When $N$ is Double \label{appendix:results_double_n}}

We hypothesize that the information a rationale holds, increases with increasing rationale lengths similar to \citet{jain2020learning}. We therefore evaluate the effectiveness of our approach, when doubling the upper-bound of the maximum allowed rationale length $N$ (see \S\ref{sec:methodology}). We assume that this should result into better rationale comprehensiveness and sufficiency.

In Table \ref{tab:rationale_ratios_doble} we present the computed rationale lengths when we double $N$. As we observe our rationales are still shorter compared to $N \times 2$, with certain cases resulting in significant reductions. For example in Ev.Inf. with $\alpha$, contiguous rationales extracted with our approach are on average 7.5\% shorter than fixed length rationales (approximately 27 tokens shorter).

\begin{figure}[!t]
    \centering
    \begin{subfigure}[b]{0.15\textwidth}
         \centering
         \includegraphics[width=\textwidth]{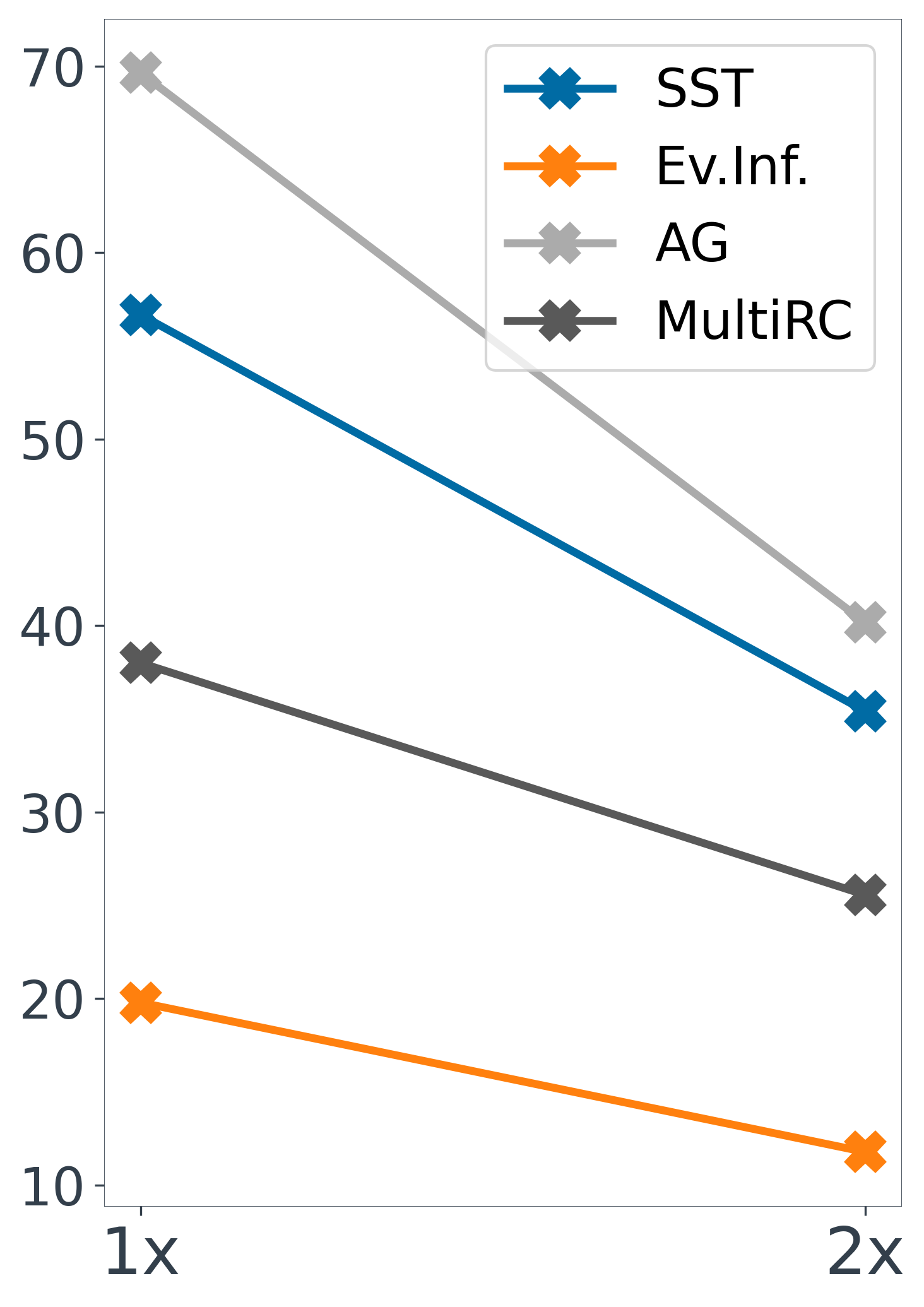}
         \caption{F1}
     \end{subfigure}
     \hspace{0.0em}
     \begin{subfigure}[b]{0.15\textwidth}
         \centering
         \includegraphics[width=\textwidth]{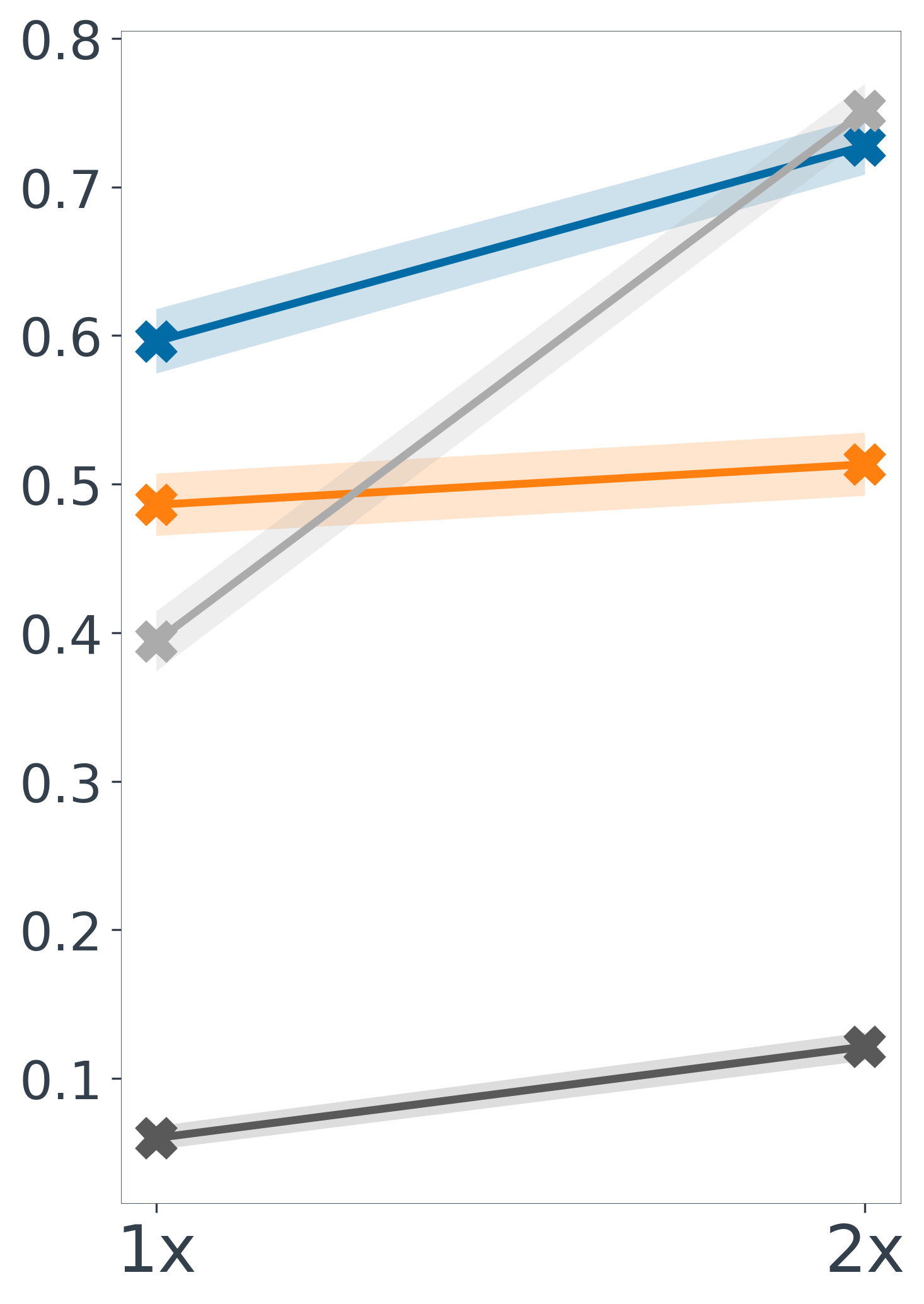}
         \caption{NormSuff}
     \end{subfigure}
     \hspace{0.0em}
     \begin{subfigure}[b]{0.15\textwidth}
         \centering
         \includegraphics[width=\textwidth]{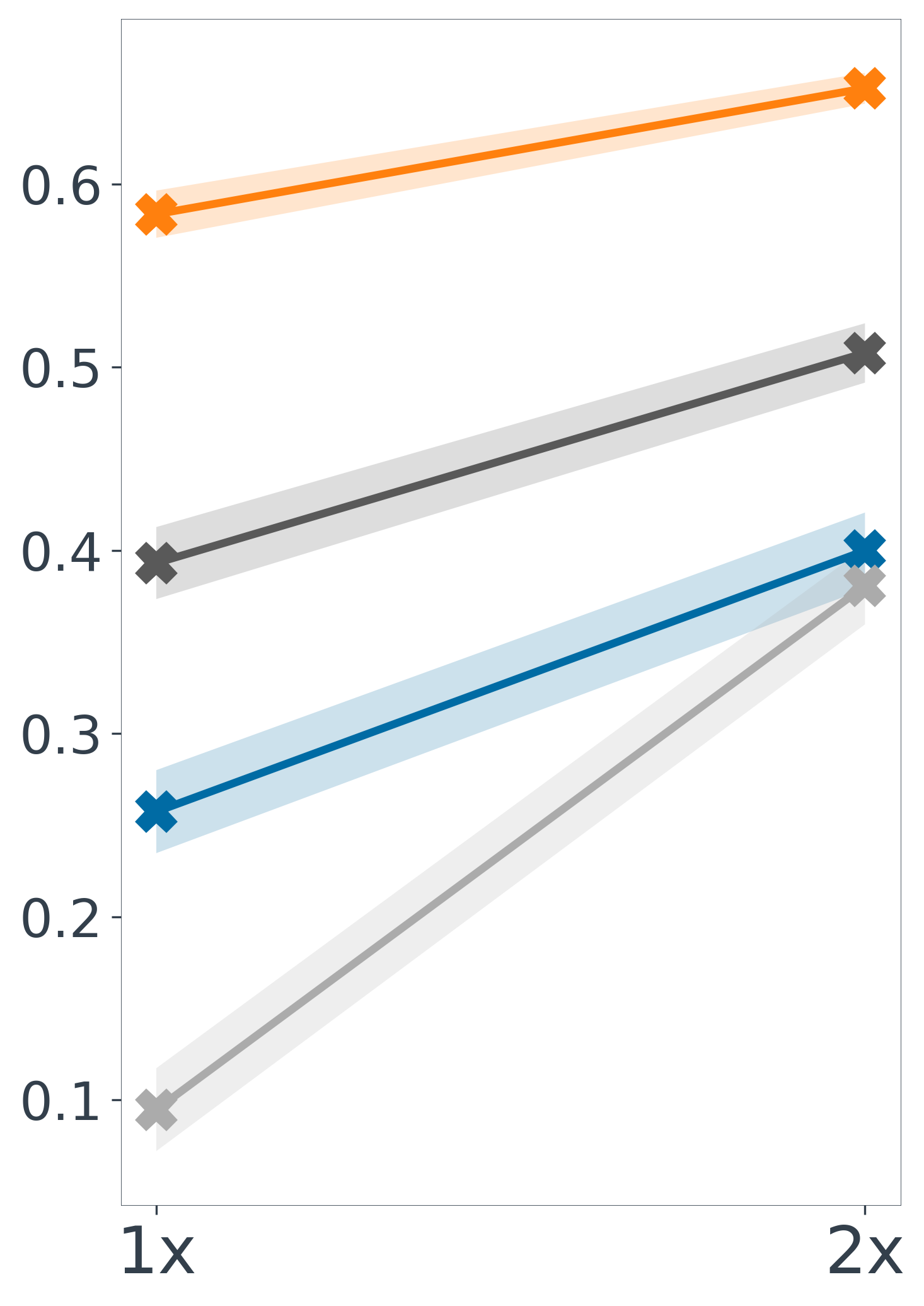}
         \caption{NormComp}
     \end{subfigure}
     \hspace{0.0em}
    \caption{F1 macro (lower is better), mean NormSuff and NormComp scores (higher is better) extracted rationales when .}
    \label{fig:double_lengths}
\end{figure}

Figure \ref{fig:double_lengths} shows F1 macro, NormSuff and NormComp scores of rationales, when we increase $N$ (1x) to $2 \times N$ (2x). For brevity we show results from rationales extracted with our best performing method (i.e. using instance-level feature scoring method, rationale length and type). Results highlight that our approach successfully scales with increasing rationale lengths, resulting in more sufficient and comprehensive rationales with average length shorter than the $2 \times N$ upper-bound.

\renewcommand*{\arraystretch}{1}
\begin{table}[!t]
\small
\centering
\setlength{\tabcolsep}{2pt}
\begin{tabular}{cl||cccccc|c}
{} & {} &   $\mathbf{x}\nabla\mathbf{x}$ &    IG & DeepLift &  LIME &     $\alpha$ &   $\alpha\nabla\alpha$ &  Avg. \\ \hline \hline
\parbox[t]{3mm}{\multirow{4}{*}{\rotatebox[origin=c]{90}{\textsc{TopK}}}} & SST    & 29.5 &  30.5 &     30.3 &  31.0 &  28.4 &  30.3 &  30.0 \\
& AG          & 32.4 &  34.8 &     32.4 &  32.8 &  33.6 &  32.9 &  33.2 \\
& Ev.Inf.       &   14.8 &  15.3 &     16.9 &  15.3 &  11.9 &  14.0 &  14.7 \\
& MultiRC       & 26.9 &  28.5 &     31.7 &  25.4 &  31.1 &  32.1 &  29.3 \\ \hline
\parbox[t]{3mm}{\multirow{4}{*}{\rotatebox[origin=c]{90}{\textsc{Cont.}}}} & SST     &  28.3 &  28.7 &     28.8 &  28.4 &  27.0 &  29.3 &  28.4 \\
& AG      &  31.7 &  32.2 &     31.8 &  30.6 &  31.4 &  31.4 &  31.5 \\
& Ev.Inf. &   13.4 &  13.6 &     15.3 &  14.4 &  12.5 &  13.0 &  13.7 \\
& MultiRC & 26.4 &  28.6 &     30.0 &  25.1 &  30.1 &  30.0 &  28.4 \\
\end{tabular}
\caption{Average instance-specific rationale lengths computed using JSD, across instances for \textsc{TopK} and \textsc{Contiguous} rationale types when we double $N$ to $2 \times N$.}
\label{tab:rationale_ratios_doble}
\end{table}

\begin{table}[!t]
    \centering
  \begin{tabular}{ll||cccc}
{} & {} & \textbf{SST} & \textbf{MultiRC} & \textbf{AG} & \textbf{Ev.Inf.} \\ \hline\hline
\parbox[t]{3mm}{\multirow{6}{*}{\rotatebox[origin=c]{90}{\textsc{TopK}}}} & \textbf{DeepLift}                         &                                  1.0 &                                      1.0 &                                 1.0 &                                      1.0 \\
& \textbf{LIME}                             &                                  1.1 &                                      0.9 &                                 1.0 &                                      1.0 \\
& $\boldsymbol{\alpha}$                     &                                  1.1 &                                      1.0 &                                 1.0 &                                      0.8 \\
& $\boldsymbol{\alpha\nabla\alpha}$         &                                  1.0 &                                      1.0 &                                 1.0 &                                      0.8 \\
& \textbf{IG}                               &                                  1.0 &                                      0.9 &                                 1.0 &                                      1.0 \\
& $\boldsymbol{\mathbf{x}\nabla\mathbf{x}}$ &                                  1.0 &                                      0.9 &                                 1.0 &                                      0.9 \\ \hline

\parbox[t]{3mm}{\multirow{6}{*}{\rotatebox[origin=c]{90}{\textsc{Contiguous}}}} & \textbf{DeepLift}                         &                                        1.0 &                                            0.9 &                                       1.0 &                                            0.9 \\
& \textbf{LIME}                             &                                        1.0 &                                            0.8 &                                       1.0 &                                            0.9 \\
& $\boldsymbol{\alpha}$                     &                                        1.0 &                                            1.0 &                                       1.0 &                                            0.8 \\
& $\boldsymbol{\alpha\nabla\alpha}$         &                                        1.0 &                                            0.9 &                                       1.0 &                                            0.8 \\
& \textbf{IG}                               &                                        1.0 &                                            0.9 &                                       1.0 &                                            0.9 \\
& $\boldsymbol{\mathbf{x}\nabla\mathbf{x}}$ &                                        1.0 &                                            0.9 &                                       1.0 &                                            0.9 \\

\end{tabular}
    \caption{Relative Improvement (R.I.) for F1 macro, when moving from fixed length rationales (see $N$ in Table \ref{tab:data_characteristics}) to rationales with instance-specific length from our proposed approach ($<$1.0 is better).}
    \label{tab:dir_of_change_appendix}
\end{table}

\subsection{Instance-specific Rationale Length}

Table \ref{tab:dir_of_change_appendix} presents the Relative Improvement (R.I.) for F1 macro, when moving from fixed length rationales (see $N$ in Table \ref{tab:data_characteristics}) to rationales with instance-specific length from our proposed approach ($<$1.0 is better).

Similarly to NormComp scores, we observe that datasets MultiRc and Ev.Inf. benefit the most from our proposed approach, resulting in relatively lower F1 macro scores irrespective of the rationale type. We consider this important, as despite of the shorter rationales, a model finds more necessary for a prediction the instance-specific length rationales extracted with our proposed method compared to the fixed-length longer ones.

\subsection{Instance-specific Feature Scoring, Length and Type 
\label{sec:varying_all_appendix}}

Table \ref{tab:var_all_appendix} shows F1 macro performance (lower is better) when we select at instance-level (I-L) a combination of: (1) the rationale length (\textsc{Len}); (2) the feature scoring method (\textsc{Feat.}); and (3) the rationale type (\textsc{Type}). For reference we also show the highest scoring fixed (\textsc{Fix}) feature scoring function, fixed rationale type and fixed and instance-specific rationale length.

Results show that we can obtain highly faithful rationales when selecting at instance level all parameters (\textsc{Feat.} +  \textsc{Len} +  \textsc{Type}) using our proposed approach. Selecting at instance-level all rationale settings results in lower F1 macro performance compared to any combination across all datasets. For example with MRc, F1 performance drops to just 38.00 when we select all parameters compared to 45.1 with the second best (7 F1 point difference). This highlights the efficacy of our approach in extracting rationales that are necessary for a model to make a prediction, without requiring any a priori assumptions about any of the rationale parameters.

\begin{table}[!t]
\setlength\tabcolsep{3pt}
\small
\centering
\begin{tabular}{lll||cccc}
  & & & \multicolumn{4}{c}{\textbf{F1 macro}}  \\ 
  \textsc{\textbf{Type}} & \textsc{\textbf{Len}} & \textsc{\textbf{Feat.}}   & \textbf{SST} & \textbf{MRc} & \textbf{AG} & \textbf{EvInf}  \\\hline
  
\parbox[t]{3mm}{\multirow{3}{*}{\rotatebox[origin=c]{90}{\textsc{TopK}}}} & \textsc{Fix} & \textsc{Fix} &  63.26 &                                     67.33 &                                78.80 &                                     32.59 \\

& \textsc{I-L} & \textsc{Fix}  &   64.41 &                                     62.74 &                                79.52 &                                     25.26 \\

& \textsc{Fix} & \textsc{I-L}   &       56.40 &                                     52.40 &                                68.80 &                                     26.10 \\

& \textsc{I-L} & \textsc{I-L} &    57.20 &                                     48.80 &                                69.90 &                                     21.70 \\ \hline

\parbox[t]{3mm}{\multirow{3}{*}{\rotatebox[origin=c]{90}{\textsc{Cont.}}}} & \textsc{Fix} & \textsc{Fix}  &       70.89 &                                           68.96 &                                      89.18 &                                           55.32 \\

& \textsc{I-L} & \textsc{Fix}   &             68.80 &                                           57.07 &                                      87.70 &                                           45.69 \\

& \textsc{Fix} & \textsc{I-L}   &    69.80 &                                           55.00 &                                      86.90 &                                           50.50 \\

& \textsc{I-L} & \textsc{I-L}    &          66.40 &                                           45.10 &                                      85.30 &                                           40.70 \\ \hline

\textsc{I-L} & \textsc{I-L} & \textsc{I-L} &    \textbf{56.70} &                                     \textbf{38.00} &                                \textbf{69.60} &                                     \textbf{19.80} \\
\end{tabular}
\caption{F1 macro when we select at instance-level (I-L) a combination of: (1) the rationale length (\textsc{Len}); (2) the feature scoring method (\textsc{Feat.}); and (3) the rationale type (\textsc{Type}). \{\textsc{Type}\}-\textsc{Fix}-\textsc{Fix} and \{\textsc{Type}\}-\textsc{I-L}-\textsc{Fix} values are from the best performing feature scoring method. \textbf{Bold} values denote the highest performing combination in column-wise (lower is better).} 
\label{tab:var_all_appendix}
\end{table}

\end{document}

%% file: table_of_examples.tex
\begin{table*}[!t]
    \centering
    \fontsize{9}{9}\selectfont
    \begin{tabular}{p{0.95\linewidth}}
        \sffamily
        \\

        
        \multicolumn{1}{l}{\textbf{Example 1}} \textbf{Data.:\textsc{AG} Id: test\_4614}\\

        \textbf{[\textsc{Fixed-Len + $\alpha$}]:} ... game last \hl{\textbf{Friday night will stand , the CFL announced yesterday. While}} a review ...
        
        \textbf{[\textsc{I-L-Len + $\alpha$} (Ours)]:} ... game last Friday night will stand , \hl{\textbf{the CFL announced yesterday}}. While a review ... \\
       
        \textbf{[Predicted Topic $||$ True Topic]:} Decreased significantly $||$ Decreased significantly  \\ \hline \hline
         
         \multicolumn{1}{l}{\textbf{Example 2}} \textbf{Data.:\textsc{Ev.Inf.} Id: 3162205\_2}\\
        \textbf{[\textsc{Fixed-Len + $\alpha\nabla\alpha$}]:} ... computed tomography ( 3D - CT \hl{\textbf{ ) scans . ABSTRACT.RESULTS : The control sides treated with an autograft showed significantly better Lenke scores than the study sides treated with $\boldsymbol\beta$ - CPP at 3 and 6 months postoperatively , but there was no difference between the two sides at 12 months}} .  The fusion .. \\ 
        
        \textbf{[\textsc{I-L-Len + $\alpha\nabla\alpha$} (Ours)]:} ... computed tomography ( 3D - CT ) scans \hl{\textbf{. ABSTRACT.RESULTS : The control sides treated with an autograft showed significantly better Lenke scores}} than the study sides treated with $\boldsymbol\beta$ - CPP at 3 ... \\
       
         
         \textbf{[Predicted Relationship $||$ True Relationship]:} Increased significantly $||$ No significant difference  \\ \hline \hline

         \multicolumn{1}{l}{\textbf{Example 3}} \textbf{Data.:\textsc{SST} Id: test\_694} \\
         
         \textbf{[\textsc{Fixed-Len + $\alpha$}]:}  ... Frontal \hl{\textbf{is the antidote for Soderbergh}} fans who think he s gone too commercial ... \\
         
         \textbf{[\textsc{I-L-Len + I-L-Feat} (Ours)]:}  ... Frontal is the antidote for Soderbergh fans who think he \hl{\textbf{s gone too commercial}} ... \\
         
         \textbf{[Predicted Sentiment $||$ True Sentiment]:} Negative $||$ Positive \\ \hline \hline
         
         \multicolumn{1}{l}{\textbf{Example 4}} \textbf{Data.:\textsc{SST} Id: test\_1039}\\
    
        \textbf{[\textsc{Fixed-Len + $\alpha$}]:} It 's just \hl{\textbf{incredibly}} dull. \\
        
        \textbf{[\textsc{I-L-Len + I-L-Feat} (Ours)]:} It 's just \hl{\textbf{incredibly dull}}. \\
    
        \textbf{[Predicted Sentiment $||$ True Sentiment]:} Negative $||$ Negative   
        
    \end{tabular}
    
    \caption{Examples when using our approach (Ours) to select at instance-level (I-L) a combination of the: (1) rationale length (\textsc{Len}); (2) feature scoring method (\textsc{Feat}) against our baseline of fixed-length rationales from a fixed feature scoring method.}
    \label{fig:annotations}
    
\end{table*}